\documentclass[onecolumn]{article}
\usepackage[left]{lineno} 

\usepackage{algorithm}
\usepackage{algpseudocode}
\usepackage{multirow}
\usepackage{array}
\usepackage{multicol}
\usepackage{amsfonts}
\usepackage{bm}
\usepackage{xcolor}
\usepackage{comment}
\usepackage{dblfloatfix}
\usepackage{newfloat}
\usepackage{listings}
\usepackage{makecell}
\usepackage{subcaption}
\usepackage[a4paper, margin=2.75cm]{geometry}

\usepackage{color}

\newcommand{\balpha}{\boldsymbol{\alpha}}
\newcommand{\bw}{\mathbf{w}}

\newcommand{\Ltrain}{\mathcal{L}_{\text{train}}}
\newcommand{\Lval}{\mathcal{L}_{\text{val}}}

\usepackage{microtype}
\usepackage{graphicx}
\usepackage{booktabs} 
\usepackage{hyperref}

\usepackage{amsmath}
\usepackage{cuted}
\usepackage{amssymb}
\usepackage{mathtools}
\usepackage{amsthm}

\usepackage[capitalize,noabbrev]{cleveref}

\theoremstyle{plain}

\theoremstyle{definition}

\theoremstyle{remark}

\usepackage[textsize=tiny]{todonotes}

\title{Architecture-Aware Minimization (A$^2$M):\\
How to Find Flat Minima in Neural Architecture Search}
\date{}

\author{
Matteo Gambella, Fabrizio Pittorino, Manuel Roveri \\
Department of Electronics, Information and Bioengineering, Politecnico di Milano \\
Via Ponzio 34/5, Milano, 20133, Italy \\
\texttt{\{matteo.gambella, fabrizio.pittorino, manuel.roveri\}@polimi.it}
}

\begin{document}

\maketitle

\begin{abstract}
Neural Architecture Search (NAS) has become an essential tool for designing effective and efficient neural networks. In this paper, we investigate the geometric properties of neural architecture spaces commonly used in differentiable NAS methods, specifically NAS-Bench-201 and Differentiable Architecture Search (DARTS). 
By introducing notions of flatness in architecture space such as neighborhoods and accuracy barriers along paths, we reveal locality and flatness characteristics analogous to the well-known properties of neural network loss landscapes in weight space. In particular, we unveil the detailed geometrical structure of the architecture search landscape by uncovering the absence of barriers between well-performing architectures, finding that highly accurate architectures cluster together in flat regions, while suboptimal architectures instead remain isolated, showing higher values of the barriers.
Building on these insights, we propose Architecture-Aware Minimization~(A$^2$M), a novel analytically derived algorithmic framework that \emph{explicitly} biases, for the first time, the gradient of differentiable NAS methods towards flat minima \emph{in architecture space}. A$^2$M consistently improves generalization over state-of-the-art DARTS-based algorithms on benchmark datasets including CIFAR-10, CIFAR-100, and ImageNet-16-120, across both NAS-Bench-201 and DARTS search spaces. Notably, A$^2$M is able to increase the test accuracy, on average across different differentiable NAS methods, by~+3.60\% on CIFAR-10,~+4.60\% on CIFAR-100, and~+3.64\% on ImageNet-16-120 - while finding architectures with low accuracy barriers. A$^2$M can be easily integrated into existing differentiable NAS frameworks, offering a versatile tool for future research and applications in automated machine learning. Our code is available at~\url{https://github.com/AI-Tech-Research-Lab/AsquaredM}.
\end{abstract}

\section*{Keywords}
Neural Architecture Search (NAS), Differentiable NAS, Sharpness Aware Minimization, Loss landscapes, Flatness

\section{Introduction}

Neural Architecture Search (NAS) has emerged as a powerful paradigm in machine learning, offering the potential to automatically identify optimal neural network (NN) architectures for a given task~\cite{nasframework}. In recent years, NAS has gained broad attention due to its versatility and applicability in scenarios where computational or hardware constraints demand efficient and specialized models, such as mobile devices or edge computing environments~\cite{cnas, HardwareNAS}.
Fundamentally, NAS can be framed as a discrete optimization process over a vast space of neural architectures. Early approaches relied on methods like genetic algorithms~\cite{eaoptimization} and reinforcement learning~\cite{rlintro}. However, the high computational cost associated with these methods motivated the development of more efficient strategies, resulting in the introduction of differentiable relaxations of the problem, such as Differentiable Architecture Search (DARTS)~\cite{liu_darts_2019, Zela2020Understanding} and its numerous variants~\cite{darts_survey}, which offer a more tractable way to navigate large architecture spaces. These methods were also promising in terms of performance, making them increasingly popular in the field.
While considerable research efforts have been devoted to understanding the geometry of neural network loss landscapes in weight space~\cite{star, shaping} and new classes of algorithms targeting these flat regions have been developed~\cite{sam, deepBP}, the precise geometry of architecture spaces remains largely underexplored~\cite{neighborhood-aware}. A deeper understanding of architecture geometry is crucial for designing more effective NAS algorithms~\cite{losslandscapes}, and for gaining insights into both the nature of the neural architecture optimization problem and the fundamental question of \emph{why certain architectures generalize better than others}~\cite{flatnas}. Additionally, architecture geometry has been shown to be relevant even for robustness to adversarial attacks \cite{RobustNAS}.

In this work, we shed light on the impact of the geometry into the generalization of the NN by focusing on two representative differentiable NAS search spaces: the NAS-Bench-201 benchmark dataset~\cite{nasbench201} and the DARTS search space~\cite{liu_darts_2019}. Our investigation aims to uncover and quantify the geometrical properties of these neural architecture spaces.
We introduce new procedures to explore neighborhoods and accuracy paths in these spaces, allowing us to systematically examine how performance changes as we traverse from one architecture to another, possibly distant one~\cite{toroids2}. 
Our analysis reveals that \emph{high-accuracy architectures tend to cluster in flat regions of the architecture space, whereas less accurate architectures appear in more isolated regions}. This behavior suggests the existence of \emph{basins of good architectures} analogous to the flat minima observed in weight space optimization~\cite{entropic1, entropic2}.

Building on these observations, we introduce Architecture-Aware Minimization (A$^2$M), an algorithmic approach that redefines the bi-level optimization process in DARTS. 
The novel contribution of A$^2$M resides in its ability to reformulate the gradient update on the \emph{architecture} space 
by explicitly shaping the optimization process in the architecture landscape towards flat regions in architecture space - as confirmed by accuracy barriers computations - and analytically extending to the architecture space the concept
of Sharpness-Aware Minimization (SAM)~\cite{sam} in weight space. 
This shift from optimizing flatness in weight space to optimizing it in architecture space improves the generalization performances of our algorithm beyond existing DARTS-based methods.
We demonstrate the effectiveness of A$^2$M across multiple benchmarks, including CIFAR-10, CIFAR-100, and ImageNet-16-120, on both NAS-Bench-201 and DARTS search spaces. In particular, A$^2$M increases test accuracy by +3.60\% on CIFAR-10, +4.60\% on CIFAR-100, and +3.64\% on ImageNet-16-120, on average over several state-of-the-art (SotA) DARTS methods, and can be easily integrated with differentiable NAS procedures.

Our novel contributions are summarized as follows:
\begin{enumerate}
\item Through the definition of distance in architecture space, we (i)~introduce the notion of architecture neighborhoods at a given radius to unveil clusters of high-performing architectures, and (ii)~we are able to construct paths between architectures at a given distance, providing deeper insights into architecture accuracy landscapes and their design properties.
\item  
We applied a SAM-like update rule for the architecture parameters in DARTS, providing the basis for the analytical derivation of our Architecture-Aware Minimization (A$^2$M) algorithm explicitly targeting \emph{flat regions in architecture space}.
\item By easily plugging this new update rule into existing DARTS-based frameworks, we achieve substantial generalization improvements in the final networks across several SotA DARTS-based methods, as demonstrated by our empirical results.
\end{enumerate}

The remainder of the paper is organized as follows. In Section~\ref{sec:literature}, we review related literature on flatness concepts in NAS. Section~\ref{sec:background} provides background on the bi-level optimization procedure of DARTS methods and on the SAM gradient step. Section~\ref{sec:4} presents the construction of neighborhoods and (shortest) paths for exploring the geometry of neural architecture spaces. In Section~\ref{sec:experiments}, we introduce the A$^2$M algorithm and offer detailed experimental results comparing it with other SotA DARTS-based methods, showing the superior performances of our method. Finally, Section~\ref{sec:conclusions} concludes the paper and outlines future research directions for integrating principled architecture loss landscape exploration in the NAS algorithmic design. 

\section{Related literature}
\label{sec:literature}

Many recent variants of Differentiable Architecture Search (DARTS) have focused on mitigating the so-called performance collapse or discretization gap, which arises when the supernet validation performance deviates substantially from the one of the final discrete architecture. 
A common cause is the supernet’s tendency to overfit by relying heavily on skip connections. 
Several methods constrain or balance skip connections to address this issue. 
DARTS$-$~\cite{darts-} adds auxiliary skip connections between nodes to ensure more uniform contributions from candidate operations. 
FairDARTS~\cite{chu2020fair} imposes a fairness constraint to penalize dominance by any single operation, while $\beta$-DARTS~\cite{betadarts} employs a bilevel optimization with an improved regularization term on the architecture parameters, subsequently refined by $\beta$-DARTS$++$~\cite{betadarts++} for greater stability and efficiency. 
Other approaches, among others, include 
DARTS+~\cite{liang2019darts+}, which terminates the search early to prevent overfitting, 
PC-DARTS~\cite{pc-darts}, which lowers memory use by sampling a subset of channels for each operation during the search, thus reducing bias toward skip connections, 
and $\Lambda$-DARTS~\cite{lambdadarts}, which introduces two new loss regularization terms that aim to prevent performance collapse by harmonizing operation selection via aligning gradients of layers. 
DARTS-PT~\cite{wang2021rethinking} instead re-examines how architectures are selected at the end of the search, where architecture parameters are chosen using the \emph{argmax} operator, and proposes a perturbation-based architecture selection scheme to measure each operation’s influence on the supernet, along with progressive tuning.

A complementary body of work tackles the discretization gap through the use of regularization strategies with the aim of enhancing architectural flatness, mainly investigating the properties of the continuous relaxation of the architecture space instead of the true, discrete one. We provide a comprehensive picture of the discrete architecture space, together with a principled and explicit way to bias the search process towards flat regions in this space. 
Robust DARTS (R-DARTS)~\cite{Zela2020Understanding} is among the first works addressing the instability issues in DARTS. This contribution highlights the role of the Hessian of the validation loss in DARTS, observing that smaller dominant eigenvalues correlate with improved stability and generalization and introducing improved regularization strategies to promote smoother loss surfaces in the search process. 
Shu et al.~\cite{Shu2020Understanding} visualize the DARTS loss surface by relying on continuous architecture parameters, and show that it can converge to sharp minima, leading to poor generalization. 
Therefore, they stress the importance of finding new methods to promote flatter regions in architectural space. 
Other studies have indirectly or empirically approached the flatness problem without relying on principled or explicit flatness optimization. 
Improvements on the original DARTS method have been observed through its combination with techniques such as self-distillation~\cite{selfdistill}, which improves generalization by transferring knowledge from teacher to student supernets, or the injection of noise into architecture parameters in Chen et al.~\cite{chen20}, that propose indirect ways of promoting flatness in DARTS, such as SDARTS (with two variants: random smoothing (SDARTS-RS) and adversarial perturbations (SDARTS-ADV)).
NA-DARTS~\cite{neighborhood-aware} modifies architecture connections (e.g., replacing a convolutional layer with a skip connection) to define a notion of proximity between architecture configurations, and aims to optimize the flatness around them by employing an objective function focused on performance across neighboring configurations. However, this algorithm is not able to obtain significant improvements in model performance. 
Recent findings~\cite{losslandscapes} indicate that greedy local search methods such as iteratively modifying top-performing architectures in small steps can be effective in finding high-accuracy architectures if the search is started in smooth loss basins.

Our work extends these ideas by providing new tools to visualize and quantify the geometry of NAS search spaces, including the definition of neighborhoods and distance in (discrete) neural architecture spaces, the introduction of paths in the loss landscape between distant architectures, and the distribution of accuracies for neighboring configurations at a certain distance (radius). A comprehensive evaluation on the NAS-Bench-201 and DARTS search spaces is conducted.
We then introduce A$^2$M, a novel class of algorithms that analytically introduces the Sharpness-Aware Minimization (SAM)~\cite{sam, usam} update step directly into the gradient of architecture parameters, \textit{explicitly optimizing flatness in the architecture space for the first time}. 
Our algorithm can be easily used in conjunction with most differentiable NAS algorithms, generally improves their performance, and offers a unified perspective on how geometric properties of the \textit{architecture space} can be leveraged to achieve improved generalization in neural architecture search.

\section{Background}
\label{sec:background}

This section provides an overview of two widely studied differentiable NAS search spaces, i.e. DARTS~\cite{liu_darts_2019} and NAS-Bench-201~\cite{nasbench201}, followed by a brief review of Sharpness-Aware Minimization (SAM)~\cite{sam} and its variant USAM~\cite{usam}. 
These concepts underlie the A$^2$M algorithm presented in Eq.~\eqref{eq:A$^2$M} and derived in Appendix~\ref{appendix:A$^2$M}. While our paper focuses on an image classification task on clean datasets, we mention other challenging benchmarks such as NAS-Bench-NLP \cite{NAS-Bench-NLP} tailored for NLP tasks and NAS-RobBench-201 \cite{RobustNAS} accounting for accuracy under adversarial attacks (i.e., with corrupted datasets). 

\subsection{DARTS}
\label{sec:darts}
DARTS is a cell-based NAS framework designed to reduce search dimensionality by concentrating on cell-level structures rather than entire models~\cite{liu_darts_2019}. 
Each cell is represented as a directed acyclic graph (DAG) with~$N$ nodes arranged sequentially and edges connecting them, where operations on each edge are drawn from a shared set of possible operations. 
Let $\mathcal{O}=\{o^{(i,j)}\}$ be the set of candidate operations between the $i$-th and $j$-th nodes. DARTS assigns continuous architecture parameters $\balpha=\{\alpha^{(i,j)}\}$, where each $\alpha^{(i,j)} \in \mathbb{R}^{|\mathcal{O}|}$ weights the choice of operation on a given edge in a weighted sum of operations, and must be optimized. 
The mixed operation is defined as
\begin{equation}
    \label{eq:mixedop}
    \bar{o}^{(i,j)}(x) \;=\; \sum_{o \,\in\, \mathcal{O}} \frac{\exp\bigl(\alpha_o^{(i,j)}\bigr)}{\sum_{o' \,\in\, \mathcal{O}} \exp\bigl(\alpha_{o'}^{(i,j)}\bigr)}\,o(x),
\end{equation}
where the weights are normalized via a softmax function and the network ultimately selects discrete operations via an \emph{argmax} over $\alpha^{(i,j)}$. 
The output of a cell is computed by applying a reduction operation, such as concatenation, to all intermediate nodes. Each intermediate node is computed based on its predecessors using the \textit{mixed operation} Eq.~\ref{eq:mixedop}. 

The search objective in DARTS is commonly posed as a bi-level optimization:
\begin{equation}
\label{eq1}
    \begin{aligned}
        \min_{\balpha}\; F(\balpha) &= \Lval\bigl(\bw^*(\balpha), \balpha\bigr),\\
        \text{s.t.}\quad \bw^*(\balpha) &= \arg\min_{\bw} \Ltrain\bigl(\bw, \balpha\bigr),
    \end{aligned}
\end{equation}
where $\Ltrain$ and $\Lval$ denote the training and validation losses, respectively. 
The parameter set $\bw$ is optimized at the lower level, while the architecture parameters $\balpha$ are optimized at the upper level\footnote{The final discrete architecture is derived by applying an \emph{argmax} operation to~$\{\alpha^{(i,j)}\}$.}. 
Differentiating $F(\balpha)$ requires the implicit function theorem~\cite{lorraine2020optimizing}:
\begin{equation}
    \label{eq:grad1}
    \nabla_{\!\balpha}F(\balpha) 
    = 
    \nabla_{\!\balpha}\Lval\bigl(\bw^*,\balpha\bigr) 
    + 
    \left(\nabla_{\!\balpha}^\top \bw^*(\balpha)\right)\,\nabla_{\!\bw}\Lval\bigl(\bw^*(\balpha), \balpha\bigr),
\end{equation}
where
\begin{equation}
    \label{eq:grad2}
    \nabla_{\!\balpha}\bw^*(\balpha) 
    = 
    -\Bigl[\nabla_{\bw\bw}^2\Ltrain\bigl(\bw^*, \balpha\bigr)\Bigr]^{-1}
    \,
    \nabla_{\balpha\bw}^2\Ltrain\bigl(\bw^*, \balpha\bigr).
\end{equation}

In practice, DARTS applies this formulation to two types of cells: \emph{normal cells}, which preserve spatial dimensions (height and width), and \emph{reduction cells}, which reduce them and increase feature depth. 
Each cell contains $4$ intermediate nodes connected by~$14$ edges, with $8$ candidate operations (including a \emph{zero} operation referring to a missing link between two nodes) on each edge~\cite{liu_darts_2019}. 
The graph includes two input nodes and one output node. In convolutional cells, the input nodes are the outputs of the previous two cells. 
After searching, the final, discrete architecture is constructed by retaining the strongest~$k$ operations (from distinct nodes) among all non-zero candidates for each intermediate node.

Each of the DARTS discretized cells allows $\prod_{k=1}^{4} \frac{(k+1)k}{2} \times (7^2) \approx 10^9$ possible DAGs without considering
graph isomorphism (recall we have 7 non-zero ops, 2 input nodes, 4 intermediate nodes with 2
predecessors each). Since there are both normal and reduction cells, the total number
of architectures is approximately $(10^9)^2 = 10^{18}$.

\subsection{NAS-Bench-201}
NAS-Bench-201~\cite{nasbench201} is an influential benchmark designed to facilitate analysis and comparison across NAS methods. 
It provides a DARTS-like search space with only normal cells consisting of $4$ internal nodes and $5$ possible operations per edge, resulting in $15{,}625$ unique architectures. The output of a cell is the output of the last internal node. Each architecture’s performance is tabulated for CIFAR-10~\cite{cifar}, CIFAR-100~\cite{cifar}, and ImageNet-16-120~\cite{imagenet16}, allowing researchers to query final accuracies without retraining from scratch.

\subsection{Sharpness-Aware Minimization in Weight Space}
\label{sec:sam_background}
Sharpness-Aware Minimization (SAM)~\cite{sam} modifies the standard stochastic gradient descent (SGD) update by encouraging convergence to flatter minima. Let $x_k$ be the parameters on which the gradient is calculated. Then the standard SGD and SAM updates can be written as
\begin{align}
    \text{SGD:}&\quad x_{k+1} = x_{k} - \eta\,\nabla f(x_{k}), \\
    \text{SAM:}&\quad x_{k+1} = x_{k} - \eta\,\nabla f\!\Bigl(x_{k} + \rho\,\tfrac{\nabla f(x_{k})}{\|\nabla f(x_{k})\|^2}\Bigr).
\end{align}
USAM~\cite{usam}, an unnormalized variant of SAM, replaces the normalized gradient perturbation with $\rho\,\nabla f(x_{k})$, yielding
\begin{equation}
    \label{eq:usam}
    \text{USAM:}\quad x_{k+1} = x_{k} - \eta\,\nabla f\!\Bigl(x_{k} + \rho\,\nabla f(x_{k})\Bigr).
\end{equation}
These updates explicitly encourage solutions that remain stable under small perturbations in parameter space, targeting optimized flatness in the NN loss landscape. 
Appendix~\ref{appendix:A$^2$M} details how we analytically introduce USAM Eq.~\eqref{eq:usam} into the DARTS bi-level framework Eqs.~\eqref{eq:grad1}~and~\eqref{eq:grad2} to derive the A$^2$M approximated gradient presented in Eq.~\eqref{eq:A$^2$M}.

Many works discuss how to improve SAM. ASAM \cite{ASAM} extends SAM by introducing directional adaptivity to the perturbation applied during training. CR-SAM \cite{CR-SAM} introduces a confidence regularization term to SAM's objective to avoid over-confident predictions. AdaSAM \cite{AdaSAM} dynamically adjusts the SAM perturbation radius ($\rho$) during training, based on local sharpness statistics. AESAM \cite{ae-sam} merges adversarial training with SAM by generating adversarial perturbations in both the input space and weight space. Finally, GA-SAM \cite{ga-sam} modifies the perturbation direction by aligning it with the loss gradient direction that generalizes better.

\section{Flatness Metrics and Algorithms in NAS}
\label{sec:4}

In this section, we introduce neighborhoods and paths in discrete NAS spaces to quantitatively investigate their geometry and flatness properties. Inspired by this investigation, we proceed to propose A$^2$M, a new differentiable NAS gradient update step that is easily deployable on top of most DARTS-based methods and improves their performances.

\subsection{NAS Space Geometry}
\label{sec:NASgeometry}

To investigate the architecture accuracy landscape of discrete NAS spaces, we introduce methods to analyze its key geometrical properties.
Our evaluation examines accuracy distributions at varying distances from a reference architecture, including short-range (radius 1) and long-range (up to radius 3) geometrical properties, and the impact of architectural modifications on model performance.
By analyzing accuracy distributions across different regions of the search space, we identify clustering patterns and flatness properties that influence generalization performances of architectures.
A$^2$M, our proposed optimization algorithm, was designed with the goal of targeting flat regions. Its application uncovers and confirms these insights into the structural properties of NAS spaces, contributing to a deeper understanding of their geometry.
\begin{algorithm}[b]
\caption{Encode DAG to Architecture Encoding in NASBench201}
\label{alg:nasbench_encoding}
\begin{algorithmic}[1]
\Require Graph $G$, Operation Set $\mathcal{O}$, Number of Nodes $n$
\Ensure Vector $C$ representing the architecture
\State Initialize empty list $C$
\State Define mapping $\text{OpIndex}[o] \gets$ index of operation $o$ in $\mathcal{O}$
\For{$i = 1$ to $n - 1$}
    \For{$j = 0$ to $i - 1$}
        \If{$(j, i) \in G$}
            \State $o \gets G[(j, i)]$
            \State $c \gets \text{OpIndex}[o]$
            \State Append $c$ to $C$
        \EndIf
    \EndFor
\EndFor
\State \Return $C$
\end{algorithmic}
\end{algorithm}
\begin{figure}[t]
    \centering
    \includegraphics[width=0.8\columnwidth]{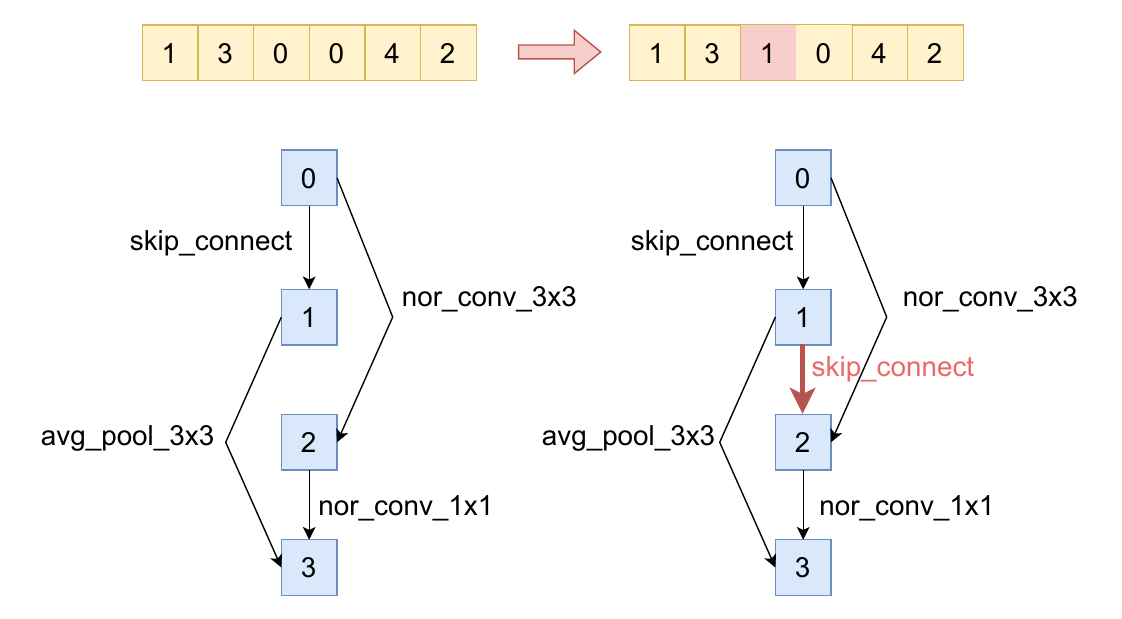} 
    \caption{Example of a NAS-Bench-201 architecture and its radius-1 neighbor. The upper part shows the vector encoding of a NAS-Bench-201 architecture (left) and its radius-1 neighbor (right). 
    The set of operations corresponding to the indexes of $\mathcal{O}$ are \{\texttt{0: none}, \texttt{1: skip\_connect}, \texttt{2: nor\_conv\_1x1}, \texttt{3: nor\_conv\_3x3}, \texttt{4: avg\_pool3x3}\} with \texttt{nor\_conv} referring to the standard (normal) convolution with dense connections.
    The lower part shows the graph of the NAS-Bench-201 architecture (left) and one of its radius-1 neighbor (right). The neighbor architecture is obtained by adding a new connection with a random possible operation from node 1 to node 2. This corresponds to updating the $3^{rd}$ element in the vector encoding.}
    \label{fig:nasbenchvector}
\end{figure}
\begin{figure}[h!]
    \centering
    \includegraphics[width=0.6\columnwidth]{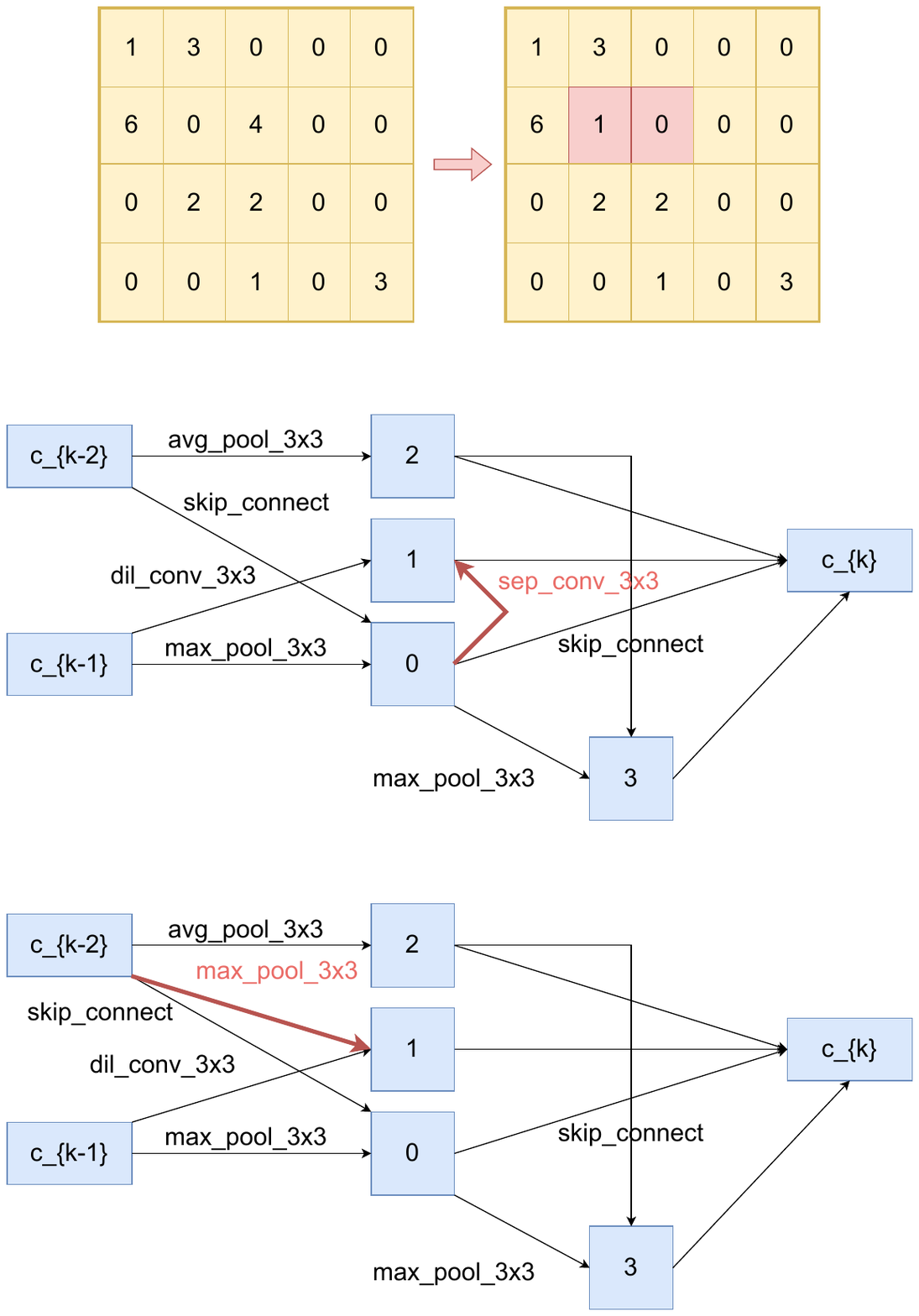} 
    \caption{Example of a DARTS architecture and one of its radius-1 neighbor. The upper part shows the adjacency matrices of the DARTS architecture (left) and its radius-1 neighbor (right). 
    The set of operations corresponding to the indexes of $\mathcal{O}$ are \{\texttt{0: none}, \texttt{1: max\_pool3x3}, \texttt{2: avg\_pool3x3}, \texttt{3: skip\_connect}, \texttt{4: sep\_conv\_3x3}, \texttt{5: sep\_conv5x5}, \texttt{6: dil\_conv\_3x3}, \texttt{7: dil\_conv5x5}\} with \texttt{dil\_conv} referring to dilated convolution and \texttt{sep\_conv} referring to separable convolution.
    The two lower graphs represent the DARTS architecture (up) and its radius-1 neighbor (down) corresponding to their adjacency matrices. The neighbor architecture is obtained by removing the connection from intermediate node 0 to node 1 and adding a new connection from the input node $c_{k-2}$ to intermediate node 1. This corresponds to updating the elements in positions $(1,1)$ and $(1,2)$ in the matrix.}
    \label{fig:adjmatrix}
\end{figure}

\begin{figure*}[h!]
    \centering
    \includegraphics[width=1.0\columnwidth]{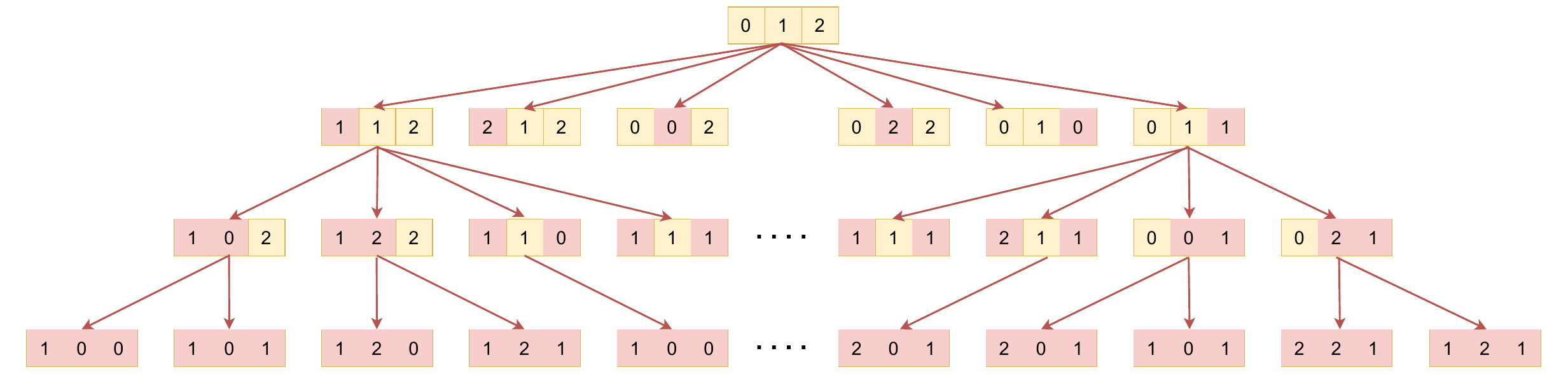}     \includegraphics[width=0.7\columnwidth]{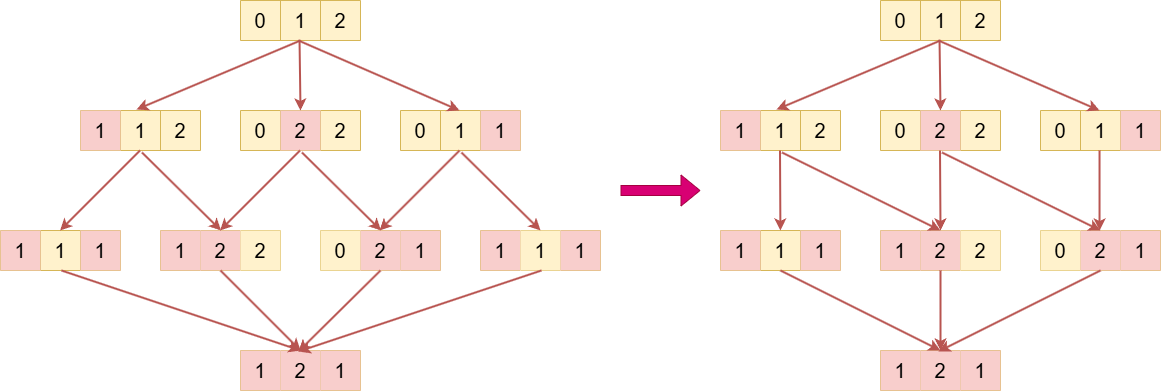} 
    \caption{(Up) Visualization of a \textit{neighbor tree} illustrating the neighborhood relationships for sequences of length 3 over a set of operations of length 3 and whose root is the initial configuration $[0,1,2]$.
    (Down) Visualization of a \textit{path tree} with a sequence of length 3 and a set of operations of length 3 from $[0,1,2]$ to $[1,2,1]$ - i.e. two architectures at maximal radius. The duplicates are counted once (see the tree on the right of the arrow).
    }
    \label{fig:neighbortree}
\end{figure*}

\subsubsection{Architecture neighborhood definition}
\label{subsec:neighborhood}

Cell-based neural architectures can be represented as sequences of length $L$ of natural numbers, where each number corresponds to a specific operation.
We define the set of possible operations as $\mathcal{O}=\{1,\,\dots,\,k\}$, where~$k$ is the total number of admissible operations. An architecture can be expressed as a sequence ($c_1, c_2, \dots, c_L$), where each~$c_i$ is an element of the set~$\mathcal{O}$.
To define architecture neighborhoods within NAS-Bench-201 and DARTS search spaces, we formalize an approach based on minimal modifications to these sequences.
We define the radius $\mathcal{R}$ in architecture space as the number of minimal modifications applied to distinct elements of an architecture, and a radius-$\mathcal{R}$ neighbor as an architecture obtained by performing~$\mathcal{R}$ such modifications starting from an initial configuration.
In NAS-Bench-201, each NN architecture is represented by one type of cell, i.e. only one sequence. The sequence is obtained by encoding the DAG representation as shown in Algorithm~\ref{alg:nasbench_encoding}.

A radius-1 neighbor is obtained by modifying a single integer in the sequence, effectively altering one operation in the cell, as illustrated in Fig. \ref{fig:nasbenchvector}. 
In contrast, in DARTS cells are represented using adjacency matrices with dimensionality the number of intermediate nodes M and M + the number of input nodes - 1. The element at position~$(i, j)$ in the matrix ($i$ refers to rows and $j$ to columns) corresponds to the operation connecting the node~$j$ to intermediate node~$i$, with a zero value indicating that there is no connection. We highlight that $c_{k-1}$ refers to an input node, i.e., the output of the previous cell. $j=0$ refers to the connections from $c_{k-1}$, $j=1$ refers to the connections from $c_{k-2}$, and $j=i+2$, for~$i = 0, \dots, M-2$ refers to the connections from the intermediate nodes. 
To generate a radius-1 neighbor, we apply a minimal modification to the adjacency matrix, which can be one of the following two possible actions: (i)~Operation Change: Replacing an operation in an existing edge; (ii)~Connection Adjustment: Removing an edge to a node and adding a new connection with a randomly selected operation, ensuring each node maintains exactly two predecessors.
A radius-1 neighbor is thus obtained by making a single, minimal change, either by modifying an operation or by adjusting connectivity. An illustration of the process (ii), Connection Adjustment, is provided in Fig.~\ref{fig:adjmatrix}. 
We emphasize that, even if this action (Connection Adjustment) results in a Hamming distance of 2 in the adjacency matrix representation (due to one edge removal and one edge addition), Connection Adjustment is considered a single atomic modification (i.e., resulting in an architectural distance of 1 for our neighborhood definition). This is because neither deleting an edge only nor adding an edge only are permissible atomic operations if they would violate the strict constraint that each intermediate node in the DARTS cell must have exactly two predecessors. Thus, Connection Adjustment represents the minimal valid change to connectivity involving distinct nodes while maintaining cell validity.
It is relevant to acknowledge an inherent ambiguity in this definition of architectural distance. While 'Connection Adjustment' is treated as a single, radius-1 atomic operation due to the cell validity constraints, it does involve modifications to multiple elements in the underlying adjacency matrix representation, contrasting with an 'Operation Change' that might alter only a single element. From a strict representational perspective (e.g., Hamming distance), these two types of atomic radius-1 modifications would have different distance. Our current definition prioritizes the count of minimal valid state transitions. 

The equivalent integer sequence representation of the resulting perturbed adjacency matrix can be recovered by extracting the meaningful elements where~$j - i \leq 1$ for~$i = 0, \dots, M-1$. Since DARTS architectures consist of both a normal cell and a reduction cell (each represented by a distinct adjacency matrix), modifications are applied randomly to one of the two matrices.
Using this architecture neighborhood definition, we can construct a \emph{neighbor tree}, where level~$\mathcal{R}$ consists of all architectures obtained through~$\mathcal{R}$ distinct modifications from an initial architecture (i.e., all architectures at radius~$\mathcal{R}$ from the root). Modifications that reverse previous changes are not allowed. 
An example of a neighbor tree is shown in the upper part of Fig.~\ref{fig:neighbortree}.

\subsubsection{Paths in Neural Architecture Space}
\label{subsec:pathprocedure}

We introduce a methodology for constructing and analyzing paths between any two specific NN architectures, denoted $A_{1}$ and $A_{2}$. This involves generating a \emph{path tree} (illustrated in Fig.~\ref{fig:neighbortree}, lower panel) which comprehensively represents \textit{all shortest sequences of modifications} that transform $A_{1}$ into $A_{2}$. 
Each level in this path tree corresponds to architectures reachable by a specific number of modifications from the source architecture ($A_{1}$) along any shortest path towards the target ($A_{2}$). Critically, this process is distinct from an omnidirectional neighborhood expansion (as detailed in Section~\ref{subsec:neighborhood} and as can be observed by comparing the upper and lower panels of Fig.~\ref{fig:neighbortree}). Here, modifications at each step are \textit{directed}, meaning they are constrained to those that reduce the "distance" to the target architecture. For instance, in NAS-Bench-201, an action involves updating a single element in the current configuration to match the corresponding element in the target configuration.
This process can be seen as a constrained variant of the neighbor tree, introduced in Section~\ref{subsec:neighborhood}, where the set of possible moves is restricted to those that directly transition toward the target architecture.

From this path tree, we derive an \textit{accuracy path}. This path plots the accuracies of $A_{1}$ and $A_{2}$ at its endpoints. For intermediate levels, it displays the average accuracy of all \textit{unique} architectures encountered at that specific modification distance along any of the shortest paths. By averaging over only unique architectures, we ensure that frequently occurring intermediate states are not overrepresented. 
Once considering only unique architectures, and due to the nature of our modification steps and shortest path definition, these paths are symmetric; the path tree structure and the set of intermediate architectures remain consistent when considering the transformation from $A_{2}$ to $A_{1}$.
This structured and symmetric approach allows for a detailed analysis of accuracy transitions and barriers within the specific landscape connecting two architectures.


\subsection{The A$^2$M Algorithm}

We introduce A$^2$M, a novel and \textit{general} algorithmic framework introducing a novel formulation of the DARTS gradient update step in the \emph{architecture} space, specifically tailored to bias the search process towards flat architecture neighborhoods. As we show in this paper, our architecture gradient step formulation \textit{can be easily applied to most DARTS-based methods}.
Our update step is analytically derived introducing the Sharpness-Aware Minimization (SAM)~\cite{sam} update step, originally formulated in weight space (and in its unnormalized variant USAM~\cite{usam}, Eq.~\eqref{eq:usam}), in the DARTS bi-level optimization problem, Eqs.~\ref{eq1}. 
In our approach, flatness
is optimized exclusively on the architecture parameters~$\boldsymbol{\alpha}$, resulting in the A$^2$M update formula:
\begin{equation}
    \label{eq:A$^2$M}
    \begin{aligned}
    \nabla_{\boldsymbol{\alpha}}\mathcal{L}_{\text{val}}&\left(\boldsymbol{w}^{*}\left(\boldsymbol{\alpha}\right),\boldsymbol{\alpha}\right) \approx \nabla_{\tilde{\boldsymbol{\alpha}}_{\xi=0}}\mathcal{L}_{\text{val}}\left(\boldsymbol{w},\tilde{\boldsymbol{\alpha}}_{\xi=0}\right)+\rho_{\alpha}\frac{\nabla_{\boldsymbol{\alpha}_{\xi=0}^{+}}\mathcal{L}_{\text{val}}\left(\boldsymbol{w},\boldsymbol{\alpha}_{\xi=0}^{+}\right)-\nabla_{\boldsymbol{\alpha}_{\xi=0}^{-}}\mathcal{L}_{\text{val}}\left(\boldsymbol{w},\boldsymbol{\alpha}_{\xi=0}^{-}\right)}{2\epsilon}\\
    &\text{with}\quad \boldsymbol{\alpha}_{\xi=0}^{\pm} = \boldsymbol{\alpha}\pm\epsilon\nabla_{\tilde{\boldsymbol{\alpha}}_{\xi=0}}\mathcal{L}_{\text{val}}\left(\boldsymbol{w},\tilde{\boldsymbol{\alpha}}_{\xi=0}\right); \quad \tilde{\boldsymbol{\alpha}}_{\xi=0} = \boldsymbol{\alpha}+\rho_{\alpha}\nabla_{\boldsymbol{\alpha}}\mathcal{L}_{\text{val}}\left(\boldsymbol{w},\boldsymbol{\alpha}\right).
    \end{aligned}
\end{equation}
Here, $\boldsymbol{\alpha}$ are the architecture parameters, $\boldsymbol{w}$ are the network weights, and $\mathcal{L}_{\text{val}}$ is the validation loss. The term $\tilde{\boldsymbol{\alpha}}_{\xi=0}$ represents the architecture parameters shifted according to the SAM principle using a neighborhood radius $\rho_{\alpha}$. The terms $\boldsymbol{\alpha}_{\xi=0}^{+}$ and $\boldsymbol{\alpha}_{\xi=0}^{-}$ are further perturbations of $\boldsymbol{\alpha}$ by a small step $\epsilon$ along the gradient direction at $\tilde{\boldsymbol{\alpha}}_{\xi=0}$, used for the finite difference approximation of the second-order term. $\epsilon$ is usually chosen to be a fixed small value, $10^{-2}$ in our case.
The second finite differences term in Eq.~\ref{eq:A$^2$M} is significant as it represents a novel curvature component within the architecture space (absent in the standard DARTS formulation), which we have analytically derived. For the complete analytical derivation of the gradient update, refer to Appendix~\ref{appendix:A$^2$M}.

This redefinition of the DARTS architecture gradient update step can be easily integrated as a plug-in into other DARTS-based algorithms, non-trivially modifying the architecture parameters update rule rather than randomly perturbing them. 
Our approach operates at a principled, flatness-enhancing level by introducing only one additional hyperparameter $\rho_{\alpha}$ related to the radius of the architecture neighborhood over which the optimization is performed.

\subsubsection{Computational Complexity of the $A^2M$ update compared with first–order DARTS}

Let $d \coloneqq |\boldsymbol{\alpha}|$ be the number of architecture parameters and
$m \coloneqq |\boldsymbol{w}|$ the number of network weights, with $m \gg d$ in practice.
Denote by $C_{\alpha} = \Theta(d)$ the cost of a single forward–backward pass
that accumulates gradients only for $\boldsymbol{\alpha}$ and by
$C_{w} = \Theta(m)$ the cost of an analogous pass that accumulates gradients
for $\boldsymbol{w}$.
First‐order DARTS requires only the single evaluation
$\nabla_{\boldsymbol{\alpha}}\mathcal{L}_{\text{val}}(\boldsymbol{w},\boldsymbol{\alpha})$. 
While both methods share the same cost for what concerns the weight parameters $\boldsymbol{w}$, in the A$^{2}$M update there are $4$ backward passes with respect to
the architectural parameters $\boldsymbol{\alpha}$:
\begin{enumerate}
\item the \emph{base} validation gradient
      $\boldsymbol{g}_{0}= \nabla_{\boldsymbol{\alpha}}\mathcal{L}_{\text{val}}(\boldsymbol{w},\boldsymbol{\alpha})$,
      used solely to construct the SAM shift
      $\tilde{\boldsymbol{\alpha}}_{\xi=0}= \boldsymbol{\alpha} + \rho_{\alpha} \boldsymbol{g}_{0}$;
\item the validation gradient at the shifted point
      $\boldsymbol{g}_{1}= \nabla_{\boldsymbol{\alpha}}\mathcal{L}_{\text{val}}(\boldsymbol{w},\tilde{\boldsymbol{\alpha}}_{\xi=0})$,
      which is required in the update rule and provides the direction for the
      finite difference; 
\item the gradient $\nabla_{\boldsymbol{\alpha}}\mathcal{L}_{\text{val}}(\boldsymbol{w},\boldsymbol{\alpha}_{\xi=0}^{+})$
      evaluated at $\boldsymbol{\alpha}_{\xi=0}^{+}= \boldsymbol{\alpha} + \epsilon \boldsymbol{g}_{1}$; 
\item the gradient $\nabla_{\boldsymbol{\alpha}}\mathcal{L}_{\text{val}}(\boldsymbol{w},\boldsymbol{\alpha}_{\xi=0}^{-})$
      evaluated at $\boldsymbol{\alpha}_{\xi=0}^{-}= \boldsymbol{\alpha} - \epsilon \boldsymbol{g}_{1}$.
\end{enumerate}
No additional backward passes are necessary because 
$\boldsymbol{g}_{1}$ is
employed both in the update formula and to build the finite‐difference pair. 
We report a summarized comparison of the total computational for both first-order DARTS and A$^2$M in Table~\ref{tab:cost}.
\begin{table}[h]
\centering
    \begin{tabular}{lccc}
    \toprule
    Algorithm & $\#$ back‐prop.\ w.r.t.\ $\boldsymbol{w}$ & $\#$ back‐prop.\ w.r.t.\ $\boldsymbol{\alpha}$ & Leading cost \\
    \midrule
    DARTS ($1^{st}$)
    & $1$ & $1$ & $C_{w} + C_{\alpha}$ \\[2pt]
    A$^{2}$-DARTS ($1^{st}$) (Eq.~\eqref{eq:A$^2$M})
    & $1$ & $4$ & $C_{w} + 4\,C_{\alpha}$ \\
    \bottomrule
    \end{tabular}
    \caption{Comparison of the computational cost of A$^2$M compared to first-order DARTS.}
    \label{tab:cost}
\end{table}

Both methods scale linearly with $d$ in the architecture block.
A$^{2}$M multiplies the $\boldsymbol{\alpha}$ computation by a constant factor
of $4$ while leaving the weight‐side work identical to that of first‐order
DARTS.  Because $d \ll m$, the overall wall‐time increase introduced by A$^{2}$M is modest, resulting in our experiments in a $1.1 \times $ to $1.3 \times $ increase in computation time on a A40 GPU.

Enhancing the computational efficiency and reducing the convergence time of A$^2$M is a primary goal for future research. This could involve optimizing A$^2$M gradient update step, potentially by incorporating strategies from the numerous SAM algorithm enhancements that have been specifically designed for faster convergence and reduced overhead~\cite{ASAM, ga-sam}. Such optimizations would be particularly beneficial when scaling to even larger and more complex search spaces. In this context, we speculate that exploring vast DARTS search spaces, such as those presented by Zhang et al.~\cite{bigspacedarts}, could be a fruitful direction. In such landscapes, flatness-biasing algorithms like A$^2$M might offer significant advantages by more effectively navigating towards and identifying rare, robust, high-performing flat regions, which could be easily missed by methods without an explicit bias towards flatness.

\section{Experimental Results} 
\label{sec:experiments}
In this section, we present our experimental investigation, aiming to elucidate the geometrical properties of the NAS-Bench-201 and DARTS search spaces and to evaluate the effectiveness of A$^2$M in enhancing DARTS-based methods. We emphasize that $A^2M$ can be applied to every other differentiable NAS method, but it could be extended to other NAS methods in our future work, since even discrete methods could benefit from exploiting flatness. Our procedures to analyze the geometrical properties can be extended to a broader range of search spaces, since only an encoding of the architecture is required.
We evaluate our methods on three benchmark datasets: CIFAR-10~\cite{cifar}, CIFAR-100~\cite{cifar}, and ImageNet-16-120.
CIFAR-10 and CIFAR-100~\cite{cifar} consist of 32$\times$32 color images, representing 10 and 100 classes, respectively. Each dataset contains 60,000 images, split into 50,000 training images and 10,000 test images. ImageNet-16-120~\cite{imagenet16} is a downsized version of the ImageNet dataset specifically designed for NAS evaluation, consisting of 16$\times$16 color images spanning 120 object classes. It contains 151,700 training images and 6,000 test images.
\begin{figure*}[h]
    \centering
    \includegraphics[width=0.325\columnwidth]{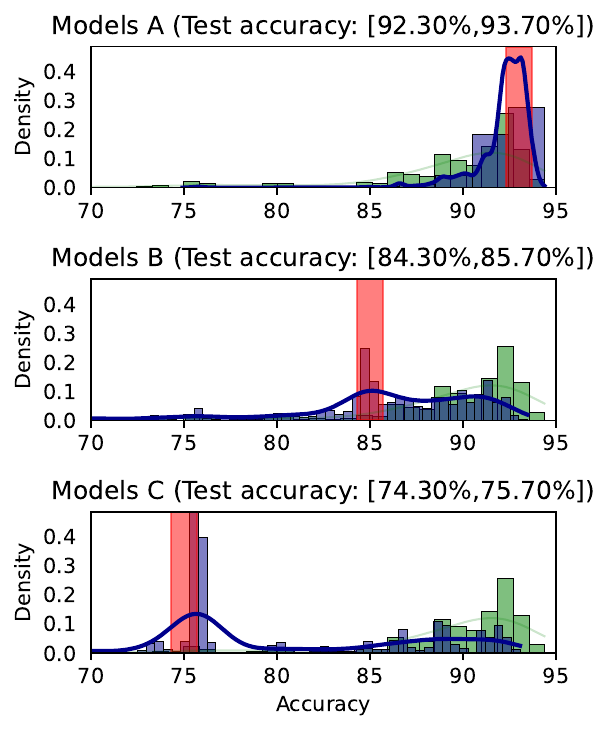} 
    \includegraphics[width=0.325\columnwidth]{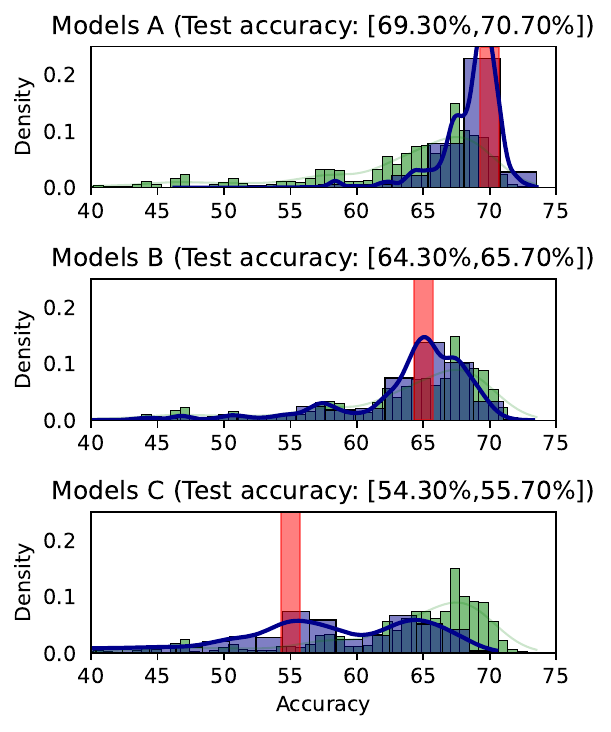} 
    \includegraphics[width=0.325\columnwidth]{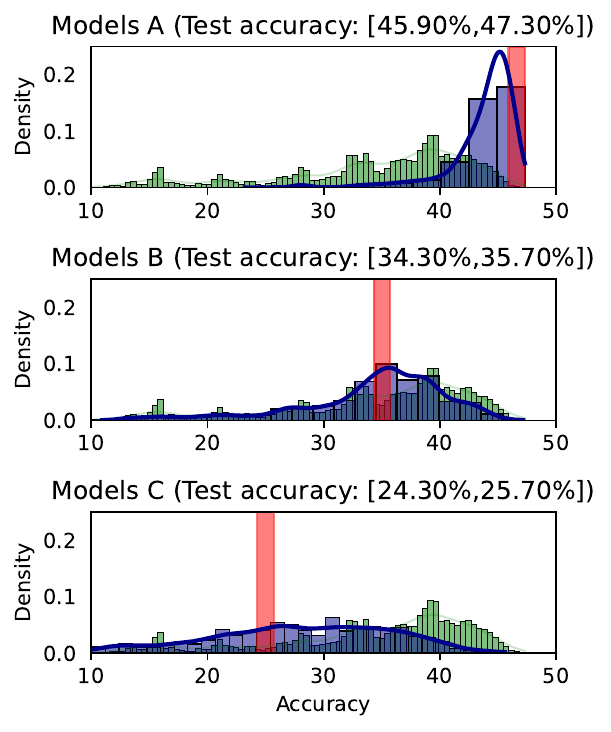} 
    \caption{
    Density histograms of test accuracies over radius-1 neighborhoods on NAS-Bench-201 and on CIFAR-10 (left), CIFAR-100 (middle) and ImageNet-16-120 (right), for different accuracy ranges of reference architectures in the search space. The red shaded area refers to the range of test accuracies of the reference architectures (also reported in each subplot title). For each dataset, three accuracy ranges for the reference architecture were identified (corresponding to high, medium, and low performance), according to the difficulty of the dataset. The green distribution refers to the accuracies over the whole search space, while the blue one refers to the accuracies over the radius-1 neighborhoods only.
    The blue histograms reveal that radius-1 neighborhoods tend to have similar accuracies to their reference architectures, independently from
    the evident bias of the search space towards well-performing architectures revealed by the green histograms (see Appendix~\ref{statanalysis} for Kolomogorov-Smirnov tests quantitatively confirming this observation). Architectures with similar accuracies tend to geometrically cluster together, and there exist flat basins of architectures in the accuracy landscape.
    }
    \label{fig:nasbench-histograms-radius1}
\end{figure*}

\subsection{Investigation of Local Properties and Flatness in Architecture Space}

To shed light on the structural properties of the NAS-Bench-201 search space, we analyze the distribution of accuracies of architectures that are radius-1 neighbors of a set of reference architectures within a selected accuracy range. Specifically, we consider architectures that differ by a single operation and visualize their accuracy distributions, as shown in Fig.~\ref{fig:nasbench-histograms-radius1}).
For each data set (CIFAR-10, CIFAR-100, and ImageNet-16-120), we select three reference architectures representing high, medium, and low-performance models (top to bottom panel rows in Fig.~\ref{fig:nasbench-histograms-radius1}). The shaded red area in each histogram highlights the accuracy range of the reference architectures, while the overall accuracy distribution across all NAS-Bench-201 architectures is shown in green.
These plots reveal a clear clustering effect for high-performing architectures: their neighbors tend to have similar, high accuracy values. This local continuity diminishes for medium- and low-performing architectures, whose neighbor distributions are wider and more dispersed.

\begin{figure*}[t]
    \centering
    \includegraphics[width=0.60\columnwidth]{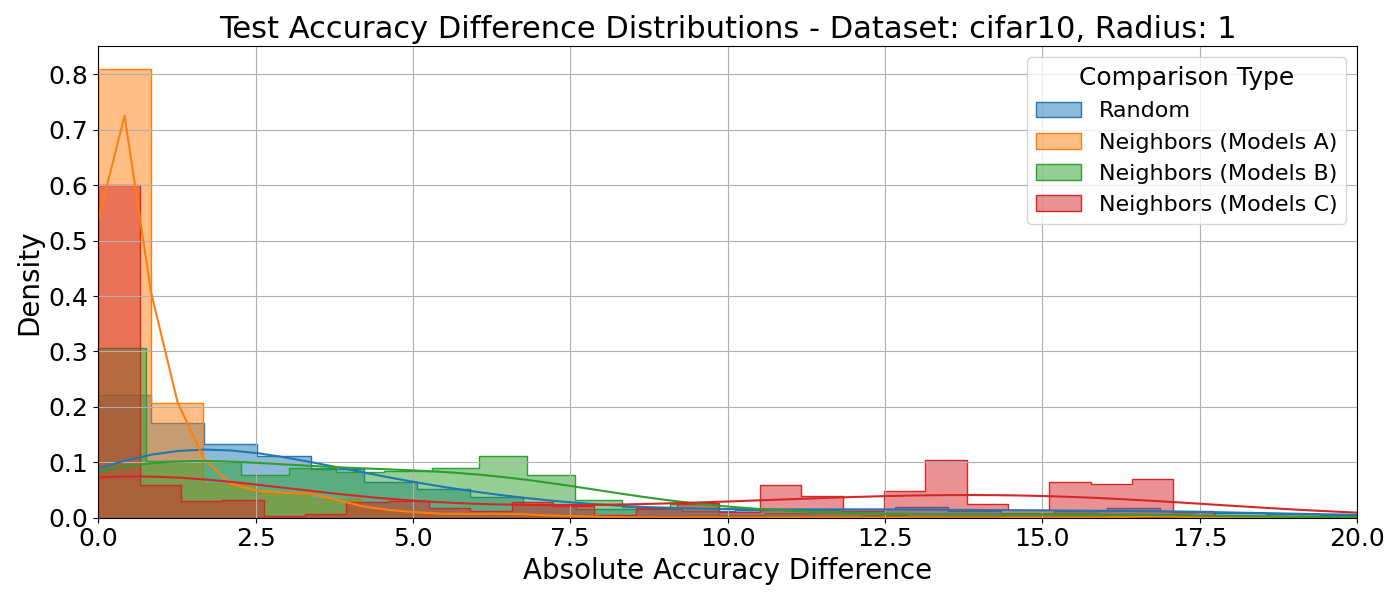} 
    \includegraphics[width=0.60\columnwidth]{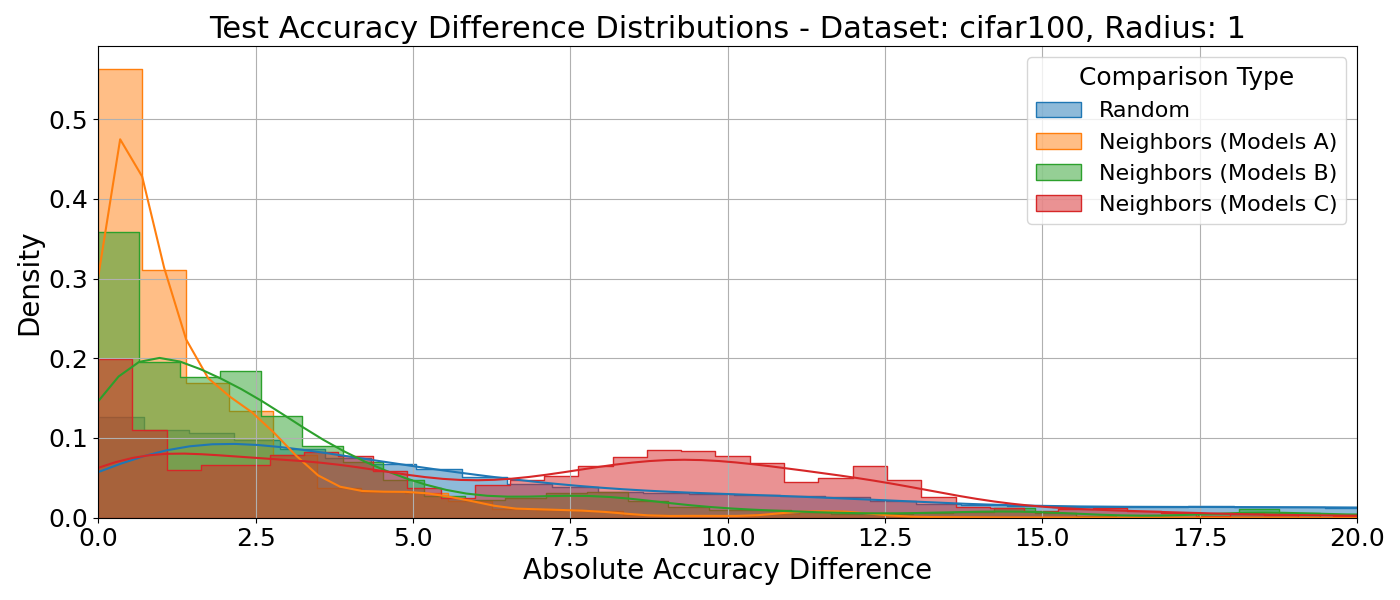} 
    \includegraphics[width=0.60\columnwidth]{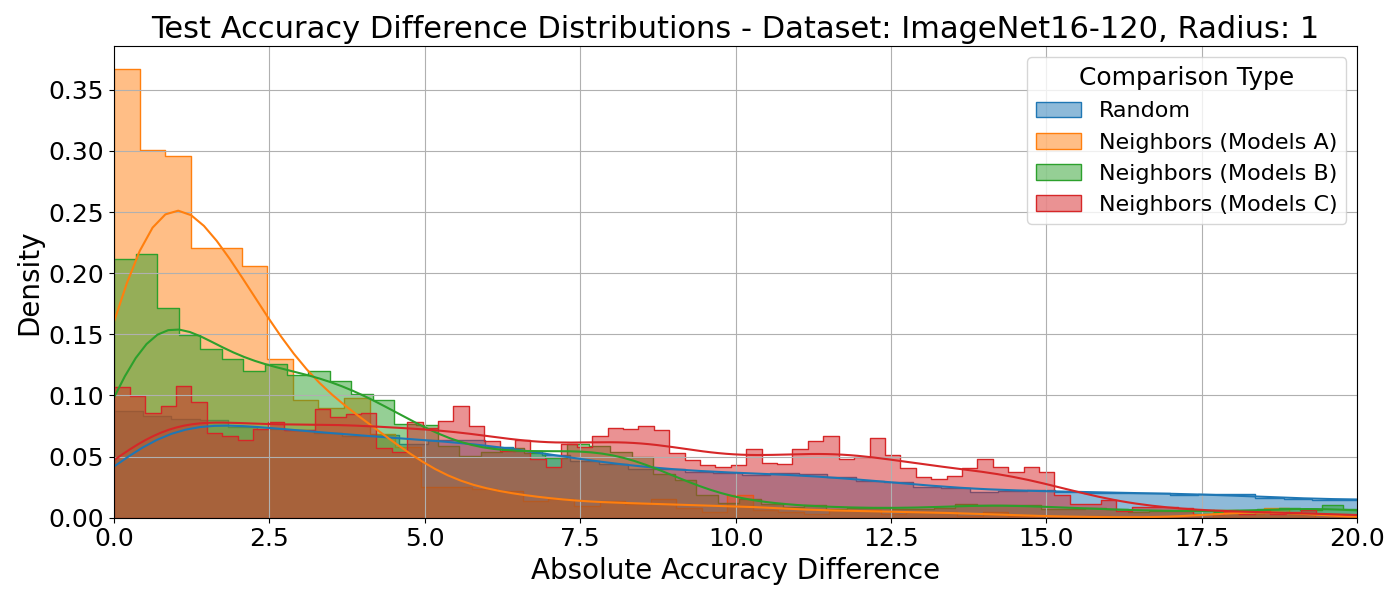} 
    \caption{
    Distributions of differences of test accuracies between random models in the search space and between reference architectures and their radius-1 neighborhoods on NAS-Bench-201 on CIFAR-10 (left), CIFAR-100 (middle), and ImageNet-16-120 (right), for different accuracy ranges of reference architectures in the search space. For each dataset, three accuracy ranges for the reference architecture were identified (corresponding to high, medium, and low performance), according to the difficulty of the dataset. The blue distribution refers to the random models, while the other ones refer to the reference architectures and their neighborhood.
    The higher density of the orange distribution on CIFAR-10 reveals the clustering property, being that models with similar accuracies tend to geometrically cluster together. 
    }
    \label{fig:diff_cluster_radius1}
\end{figure*}

To quantify this observation, we analyze the absolute test accuracy difference between reference architectures and their radius-1 neighbors, compared to differences obtained by sampling random pairs of architectures. Figure~\ref{fig:diff_cluster_radius1} presents these distributions across the three datasets. We distinguish the distributions by the performance level of the reference architectures:
(i) Orange: differences for high-performing models; (ii) Green: differences for medium-performing models; (iii) Red: differences for low-performing models; (iv) Blue: differences between randomly sampled architectures.
Across all datasets, the orange distributions are sharply peaked near zero, confirming strong clustering among high-performing architectures. This means small local changes in architecture often lead to minimal loss (or gain) in accuracy. Conversely, the green and red distributions are wider, indicating that medium- and low-performing architectures have neighbors with more variable performance. Interestingly, in some cases (e.g., CIFAR-10 mid-performing models or low-performing models on CIFAR-100 and ImageNet-16-120), the random sampling (blue) distribution shows a similar or even higher density near zero compared to the red or green distributions. This suggests that the architecture space is less structured in low- and medium-accuracy regions, where even neighboring architectures may yield significantly different performance levels. This is particularly evident on CIFAR-10, where the percentage of high-performing architectures in the search space is very high. For completeness, we report the plots with neighborhoods of radius 2 and 3, respectively, in Fig. \ref{fig:diff_cluster_radius2} and \ref{fig:diff_cluster_radius3} in Appendix \ref{appendix:radii}.

Across all datasets, architectures have a consistent sequence length (i.e., the number of edges) of $L = 6$ and a set of $k = 5$  possible operations. 
Our analysis of accuracy difference distributions (Figure~\ref{fig:diff_cluster_radius1}) reveals varying degrees of clustering across different performance tiers in the NAS-Bench-201 search space. Pronounced clustering is predominantly characteristic of high-performing architectures (e.g., Model A tier, Figure~\ref{fig:diff_cluster_radius1}), where neighbors typically exhibit very small accuracy differences relative to the reference model and compared to random pairs. This strong, tight clustering in high-accuracy regions suggests a relatively flat loss landscape where small architectural changes often yield minimal performance shifts. Conversely, medium and low-performing regions (Model B and C tiers, Figure~\ref{fig:diff_cluster_radius1}) demonstrate less tight clustering, with greater performance variability among neighbors, indicating a generally less smooth landscape.
Radius-3 neighborhood accuracy distributions can become quite dispersed, sometimes appearing even less distinctly clustered than the global search space distribution, due to the fact that the global distribution contains also the more clustered radius-1 neighbors.
Additional results for radius-2 and radius-3 neighborhoods are provided in Appendix~\ref{appendix:radii}, and the results of the Kolmogorov-Smirnov test reported in the Appendix~\ref{statanalysis} show that these distributions are statistically significantly different from the accuracy distribution of the whole search space. 


\begin{figure}[t!]
    \centering
    \begin{subfigure}[t]{0.98\textwidth}
        \centering
        \includegraphics[width=\textwidth]{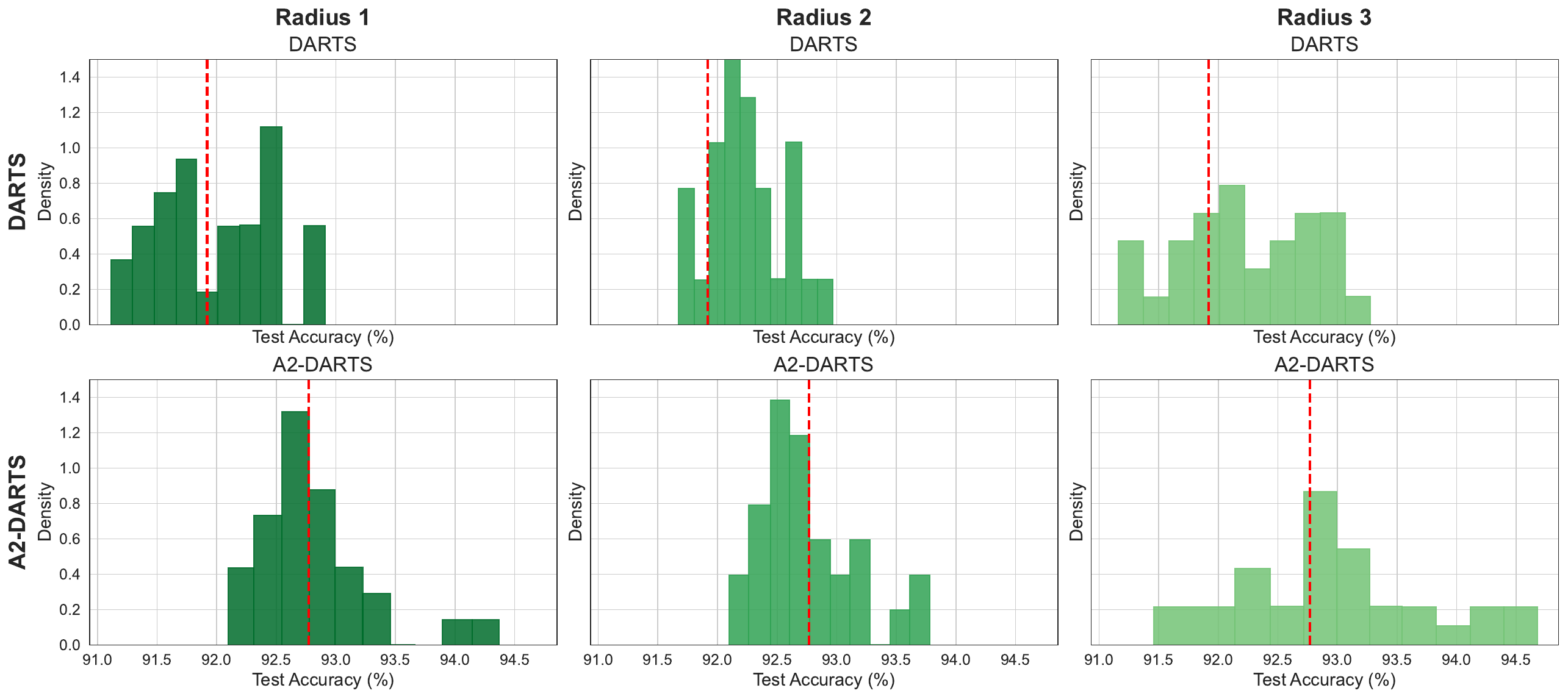}
        \caption{Test accuracy densities for neighbors of architectures found by DARTS and A$^2$-DARTS on CIFAR-10.}
        \label{fig:hist-cifar10}
    \end{subfigure}
    \vspace{1em}
    \begin{subfigure}[t]{0.98\textwidth}
        \centering
        \includegraphics[width=\textwidth]{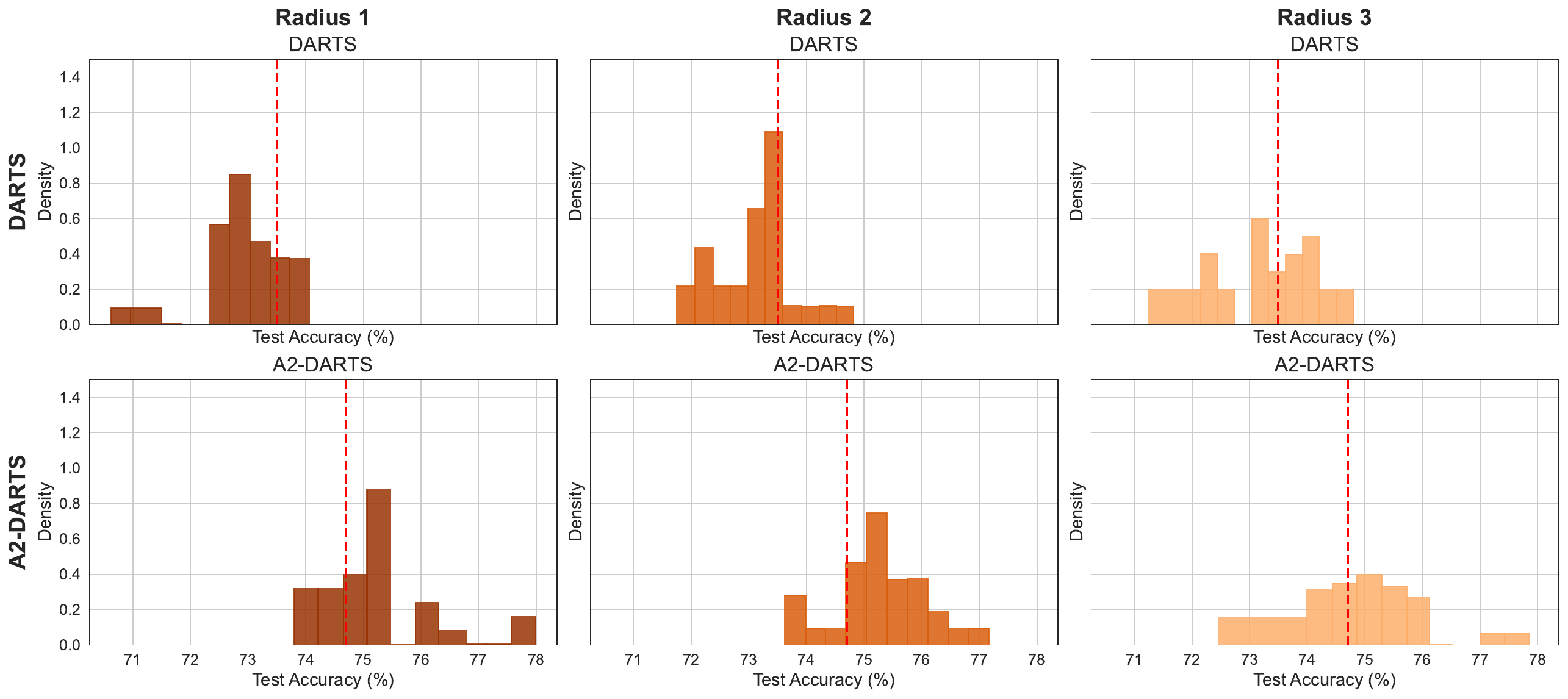}
        \caption{Test accuracy densities for neighbors of architectures found by DARTS and A$^2$-DARTS on CIFAR-100.}
        \label{fig:hist-cifar100}
    \end{subfigure}
    \caption{
        Density histograms of test accuracies for neural architectures in the neighborhood of DARTS and A$^2$-DARTS solutions, on (a)~CIFAR-10 and (b)~CIFAR-100.         
        For each dataset, 30 neighbors are sampled per radius level $\{1, 2, 3\}$ by randomly perturbing the original architecture and training them up to a fixed loss threshold. 
        Each column corresponds to a radius, while rows compare DARTS (top rows) and A$^2$M (bottom rows). The dashed red lines indicate the performance of the reference architecture. 
        We observe a local property around the initial architectures, with their neighbors generally exhibiting accuracies close to the reference, particularly for the higher-performing A$^2$-DARTS models. This tendency for high-accuracy networks and their neighbors to group together suggests the presence of relatively flat regions in the neural architecture space.
    }
    \label{fig:darts-histograms}
\end{figure}

We extend this locality analysis to the more complex DARTS search space, where $L = 14$ operations define each of the $k=8$ edges, resulting in a vastly larger, non-tabulated landscape (estimated at $\approx 10^{18}$ architectures, see Section~\ref{sec:darts}), and observe similar patterns (see Fig.~\ref{fig:darts-histograms} for the corresponding accuracy histograms on CIFAR-10 and CIFAR-100). Unlike NAS-Bench-201 where all neighbor performances can be queried directly, analyzing neighborhoods in DARTS necessitates \textit{sampling} specific architectural configurations and evaluating them through training. For this purpose, we select two neural architectures obtained by DARTS and A$^2$M respectively as reference models, and for each, we sample 30 random neighbors at radii 1, 2, and 3 using the procedure described in Section~\ref{subsec:neighborhood}. 
Notice that sampling 30 architectural neighbors in the vast DARTS space, an approach akin to finite sampling in continuous weight space studies for feasibility~\cite{Jiang2020Fantastic}, still allows for valuable insights into local landscape properties.
To ensure a consistent and fair comparison of these sampled neighbor architectures at a comparable stage of optimization, we adopted an early stopping criterion: training for each neighbor is terminated once a predefined training loss threshold is reached (0.43 for CIFAR-10 and 1.0 for CIFAR-100, typically achieved between 50-70 epochs). This methodology, which is in line with established procedures in flatness literature for exploring performance landscapes when full convergence of all samples is prohibitive~\cite{Jiang2020Fantastic}, provides a standardized evaluation point. This also makes the extensive neighborhood study for Figure~\ref{fig:darts-histograms} computationally tractable, given that training the hundreds of networks involved to full convergence 
would require hundreds of GPU days. While this strategy is viable on CIFAR-10 and CIFAR-100, it is not on ImageNet-16-120, where each architecture requires full training to reach a stable accuracy value. 

The results for DARTS on CIFAR-10 and CIFAR-100, are shown in Figure~\ref{fig:darts-histograms} for radius-1, radius-2 and radius-3 neighborhoods.
An analysis of the underlying data for the radius-1 neighborhoods in Figure~\ref{fig:darts-histograms} provides further insight into the local robustness of architectures found by A$^2$-DARTS compared to standard DARTS. Specifically, we investigate the minimum test accuracy observed among the 30 sampled radius-1 neighbors for each reference architecture. 
On CIFAR-10, the reference DARTS model achieves 91.92\% accuracy, while its least accurate sampled radius-1 neighbor performs at 91.11\% (a performance drop of 0.81\%). In contrast, the A$^2$-DARTS model, with a reference accuracy of 92.77\%, has its least accurate sampled radius-1 neighbor perform at 92.10\% (a smaller drop of 0.67\%). 
A similar and more pronounced trend is observed on CIFAR-100: the reference DARTS model (73.50\% accuracy) has a minimum radius-1 neighbor accuracy of 70.62\% (a drop of 2.88\%), whereas the A$^2$-DARTS model (74.70\% accuracy) has a minimum radius-1 neighbor accuracy of 73.80\% (a significantly smaller drop of 0.90\%). 
This quantitative comparison suggests that architectures identified by A$^2$-DARTS not only tend to reach high peak performance but also reside in immediate neighborhoods that are more robust to performance degradation from single architectural modifications. This reinforces the idea that A$^2$M helps find architectures in locally flatter and more stable regions of the landscape.
This findings reinforces the potential for gradient-based NAS methods to efficiently navigate the architecture space targeting flat regions. 

\subsubsection{Neural Architecture Paths}

\begin{figure*}[t]
    \centering        
    \begin{subfigure}[b]{0.325\columnwidth}
        \centering
        \includegraphics[width=\textwidth]{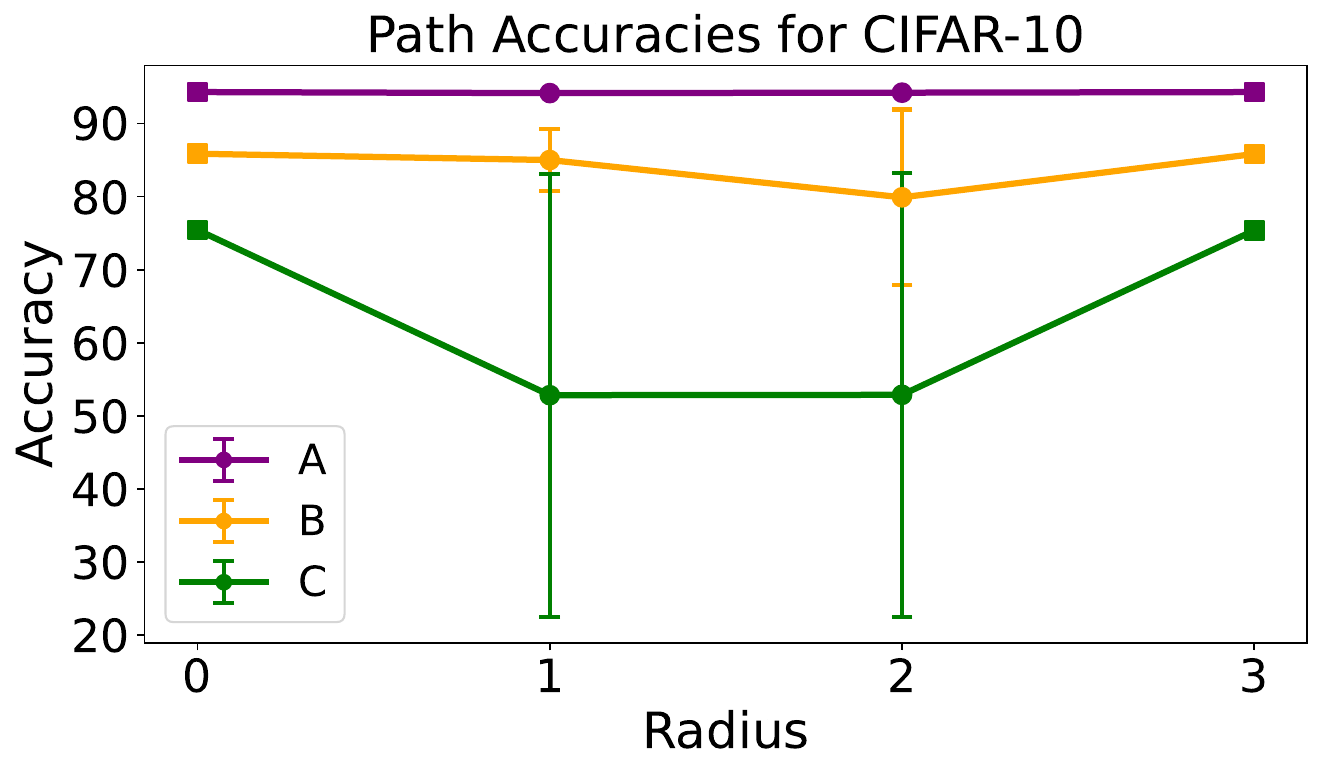} 
        \caption{CIFAR-10}
        \label{fig:paths_cifar10}
    \end{subfigure}
    \hfill 
    \begin{subfigure}[b]{0.325\columnwidth}
        \centering
        \includegraphics[width=\textwidth]{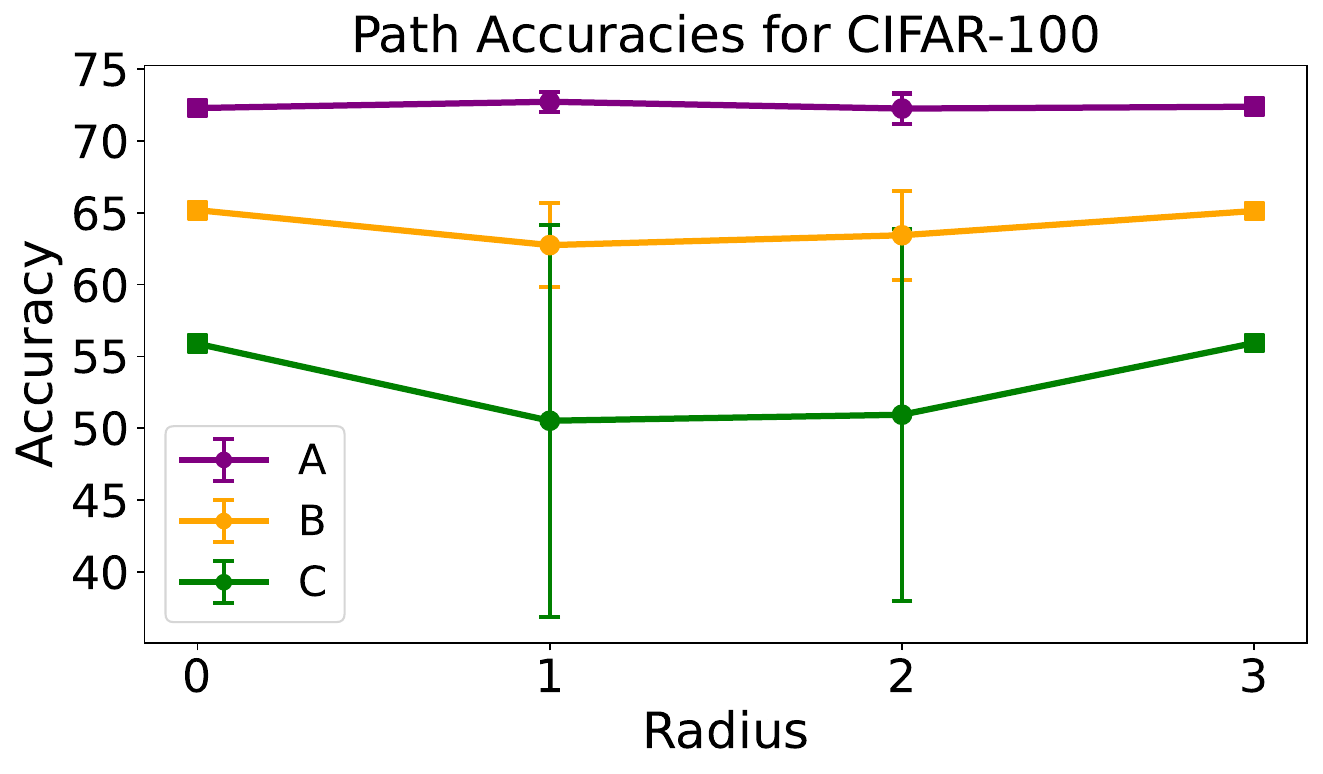} 
        \caption{CIFAR-100}
        \label{fig:paths_cifar100}
    \end{subfigure}
    \hfill 
    \begin{subfigure}[b]{0.325\columnwidth}
        \centering
        \includegraphics[width=\textwidth]{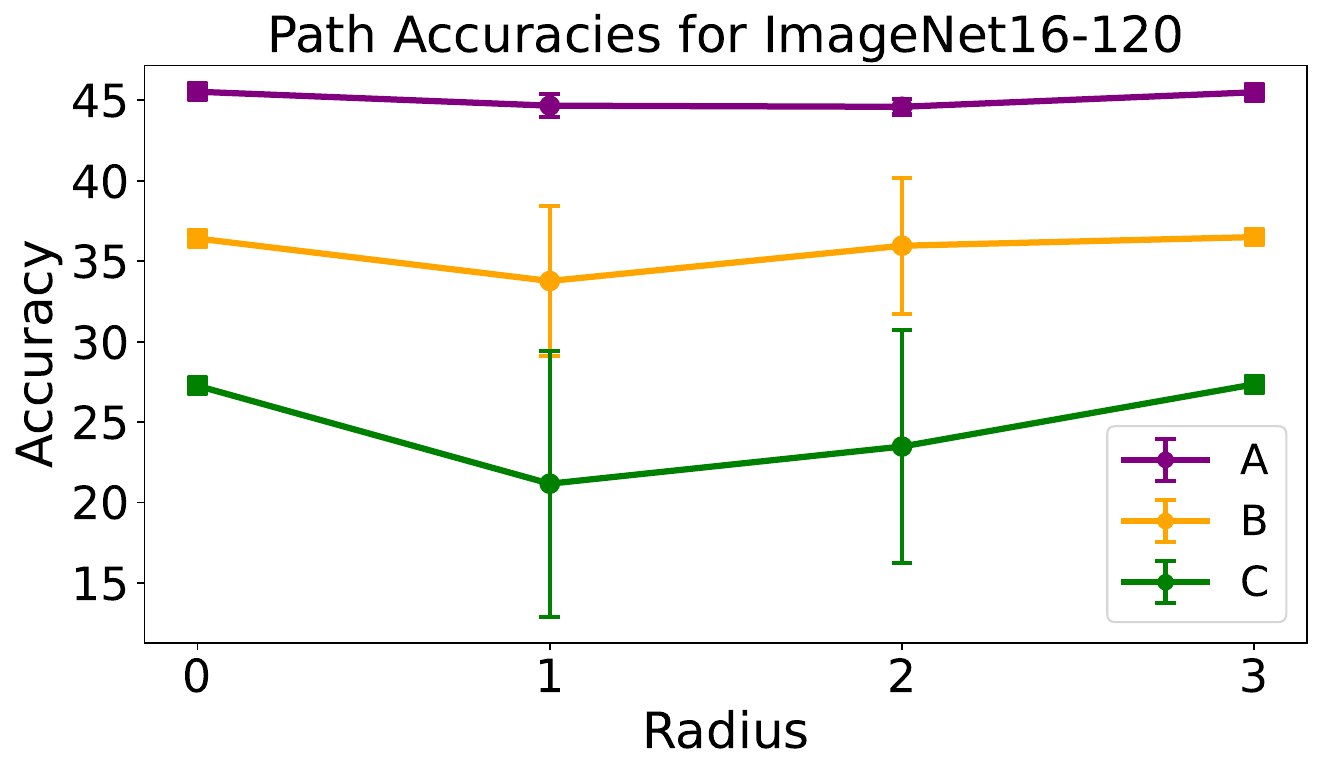} 
        \caption{ImageNet-16-120}
        \label{fig:paths_imagenet16}
    \end{subfigure}
    \caption{ 
        Paths between architectures on NAS-Bench-201 for different datasets: (a) CIFAR-10, (b) CIFAR-100, and (c) ImageNet-16-120. Each path, as defined in Section~\ref{subsec:pathprocedure}, connects a reference architecture $\mathcal{A}_1$ (high (A), medium (B), or low (C) performing, shown as separate lines in each panel; corresponds to radius 0 on the x-axis) to a distinct radius-3 neighbor $\mathcal{A}_2$ (at radius 3 on the x-axis) that has a similar initial test accuracy. The y-axis represents test accuracy, and intermediate points (radius 1, 2) show the average accuracy of unique architectures along all shortest paths. For CIFAR-10 (a), the normalized accuracy barriers for paths A, B, and C are 0.24, 22.52, and 54.14, respectively. For CIFAR-100 (b), barriers are 1.74, 4.93, and 21.56. For ImageNet-16-120 (c), barriers are 1.42, 7.11, and~10.86.     
    }
    \label{fig:paths}
\end{figure*}

For each dataset, we analyze the paths connecting architectures with three different test accuracy levels. This allows us to examine the presence of accuracy barriers between architectures with different performance levels, denoted by A, B, C in decreasing order of accuracy. 
Each path connects an architecture $\mathcal{A}_1$ to a radius-3 neighbor $\mathcal{A}_2$ with similar test accuracy, following definitions in Sec.~\ref{subsec:neighborhood}.
In Fig.~\ref{fig:paths}, we report the normalized accuracy barrier $\mathcal{B}_{\mathcal{A}_1 \leftrightarrow \mathcal{A}_2}$ between $\mathcal{A}_1$ and $\mathcal{A}_2$, defined as:
\begin{equation}
    \mathcal{B}_{\mathcal{A}_1 \leftrightarrow \mathcal{A}_2} = \frac{1}{2}\left(\text{Acc}\left(\mathcal{A}_1\right)+\text{Acc}\left(\mathcal{A}_2\right)\right)-\min_{\mathcal{A} \in \mathcal{A}_1 \leftrightarrow \mathcal{A}_2} \text{Acc}\left(\mathcal{A}\right)
\end{equation}
where $\text{Acc}(\mathcal{A})$ represents the test accuracy of an architecture~$\mathcal{A}$, and~$\mathcal{A}_1 \leftrightarrow \mathcal{A}_2$ denotes the path connecting architectures~$\mathcal{A}_1$~and~$\mathcal{A}_2$, with $\mathcal{A}$ an architecture \textit{at an intermediate step} inside this path (i.e., at radius 1 or 2 from $\mathcal{A}_1$).
Notice that, given our definition of the path tree in Section~\ref{subsec:pathprocedure}, the path between architectures $\mathcal{A}_1$~and~$\mathcal{A}_2$ is symmetric (i.e. it does not change if we start from $\mathcal{A}_1$ and we go towards $\mathcal{A}_2$ or vice-versa).

For NAS-Bench-201, we select three distinct accuracy levels for each dataset, corresponding to high (A), medium (B), and low-performing (C) models relative to the dataset’s difficulty. Our objective is to verify whether higher-performing architectures exhibit lower barriers, indicating a smoother architecture landscape.
We construct a path (see Sec.~\ref{subsec:pathprocedure}) starting at each model, and ending at a selected architecture from the search space corresponding to a radius-3 neighbor with similar accuracy. 

At each radius $n$ in the path (where $n=1,2$), we compute the average of the test accuracies of the unique $n$-radius neighbors of $\mathcal{A}_1$ that lie on a shortest path transitioning towards $\mathcal{A}_2$. The accuracies reported at the extremes (radius 0 and 3) refer to the test accuracies of $\mathcal{A}_1$ and $\mathcal{A}_2$, respectively.

The identified accuracy levels for each dataset are: 
\begin{itemize}
\item CIFAR-10 - Model~A:~94.3\%, Model~B:~85.86\%, Model~C:~75.44\%; 
\item CIFAR-100 - Model~A:~72.30\%, Model~B:~65.19\%, Model~C:~55.88\%; 
\item ImageNet-16-120 - Model~A:~45.53\%, Model~B:~36.41\%, Model~C:~27.26\%.
\end{itemize}
As shown in Fig~\ref{fig:paths}, consistently across all datasets, higher-performing architectures (paths A) exhibit the flattest paths (lowest barriers), followed by medium-performing architectures (paths B), while lower-performing ones (paths C) are characterized by the highest accuracy barriers, indicating they are more isolated. This observed ordering of barrier heights (Barrier(A) $\langle$ Barrier(B) $\langle$ Barrier(C)) strongly supports the hypothesis that flatter regions are associated with better-performing architectures.    

\begin{figure*}[t]
    \centering        
    \begin{subfigure}[b]{0.335\columnwidth} 
        \centering
        \includegraphics[width=\textwidth]{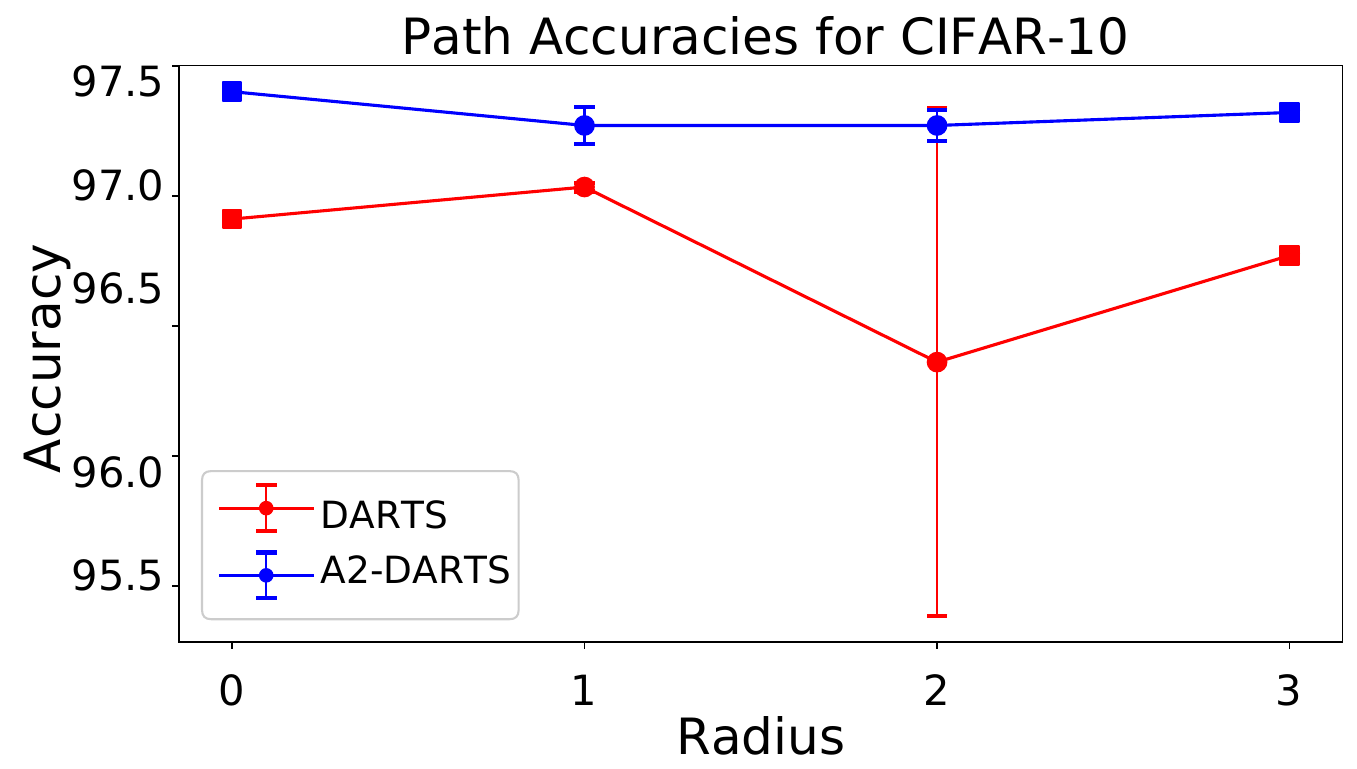} 
        \caption{CIFAR-10}
        \label{fig:pathsDARTS_cifar10}
    \end{subfigure}
    \begin{subfigure}[b]{0.325\columnwidth}
        \centering
        \includegraphics[width=\textwidth]{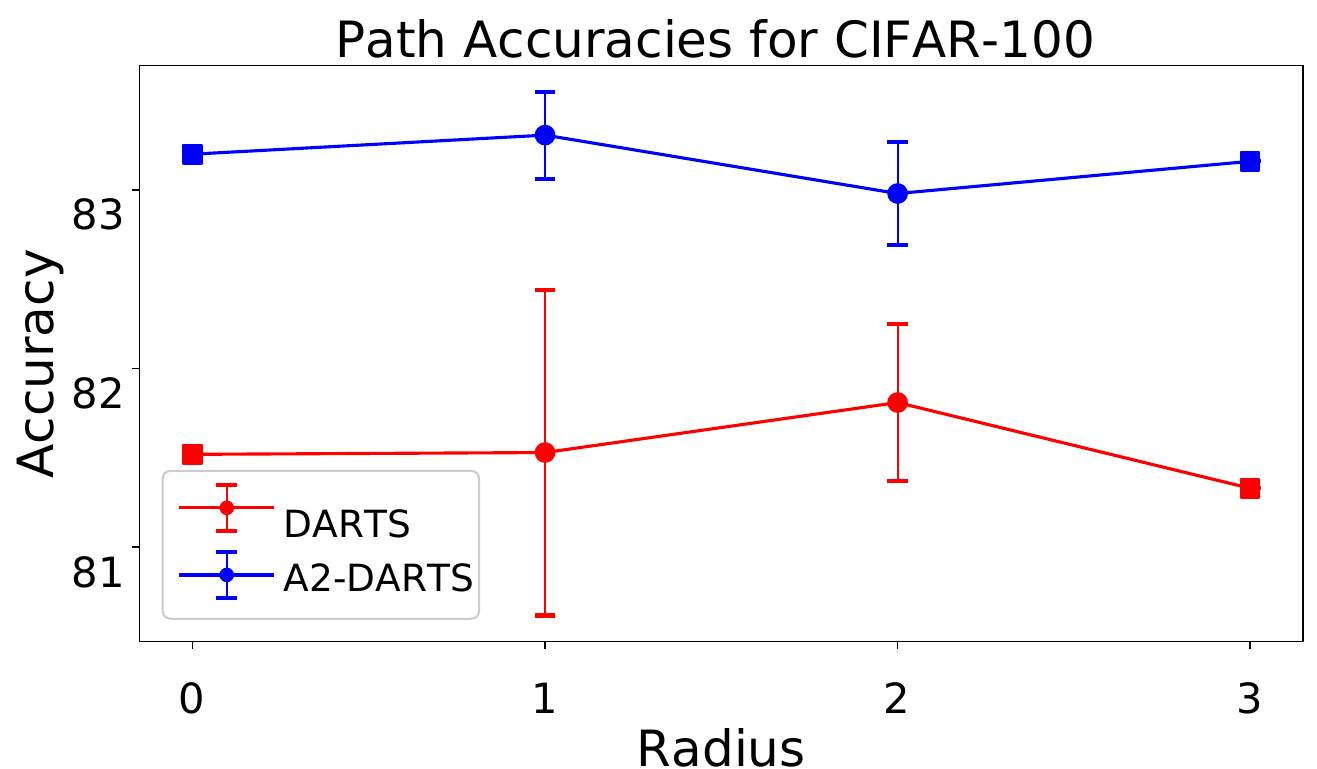} 
        \caption{CIFAR-100}
        \label{fig:pathsDARTS_cifar100}
    \end{subfigure}
    \caption{ 
    Paths between architectures found by DARTS and A$^2$-DARTS on (a) CIFAR-10 and (b) CIFAR-100. Each path, as defined in Section~\ref{subsec:pathprocedure}, connects an initial architecture $\mathcal{A}_1$ (found by either DARTS or A$^2$-DARTS; corresponds to radius 0 on the x-axis) to a distinct radius-3 neighbor $\mathcal{A}_2$ (at radius 3 on the x-axis) that was selected to have similar test accuracy. 
    The y-axis represents test accuracy, and intermediate points (radius 1, 2) show the average accuracy of unique architectures along all shortest paths. 
    The normalized accuracy barriers are: 
    for CIFAR-10 (a), DARTS: 1.99, A$^2$-DARTS: 0.15; for CIFAR-100 (b), DARTS: 1.23, A$^2$-DARTS: 0.63. 
    Architectures found by A$^2$-DARTS consistently exhibit flatter paths (lower barriers) than those found by standard DARTS.
    }
    \label{fig:pathsDARTS}
\end{figure*}

For the DARTS search space, we select two accuracy levels by considering architectures obtained from separate runs of DARTS and A$^2$M applied to pure DARTS (A$^2$-DARTS). These models were evaluated on CIFAR-10 and CIFAR-100.
As discussed in Sec.~\ref{sec:experiments}, we did not analyze ImageNet-16-120 due to its computational demands.
To compute the paths and the barriers, 
instead of selecting a predefined radius-3 neighbor, which was possible in NAS-Bench-201, we 
identify a neighbor with a similar test accuracy after partial training, truncated at a fixed train loss. This selected neighbor was then fully trained to obtain its final accuracy. All models maintained comparable accuracy levels even after full training.
The identified accuracy levels are: 
\begin{itemize}
\item CIFAR-10 - DARTS:~96.91\%, A$^2$-DARTS:~97.40\%; \item CIFAR-100 - DARTS:~81.52\%, A$^2$-DARTS:~83.20\%.
\end{itemize}
Results for the DARTS search space are reported in Fig.~\ref{fig:pathsDARTS}. 
Our findings provide strong evidence for a flatness property in the architecture landscapes of both NAS-Bench-201 and DARTS search spaces, as indicated by the systematically lower accuracy barriers observed along paths connecting well-performing architectures.

Furthermore, the \textit{geometric analysis methodology} we propose - including neighborhood definitions, path constructions, and accuracy barrier computations - is not limited to cell-based search spaces. These tools can be extended to broader NAS formulations, such as supernet-based approaches like Once-For-All (OFA)~\cite{ofa}, as they primarily require an architectural encoding and a performance evaluation mechanism. For instance, within OFA, one could define neighbors of a specific sub-network by making discrete changes to its configuration (e.g., altering kernel sizes, depth, or width at specific layers). The performance of these neighboring sub-networks can then be efficiently estimated by inheriting weights from the pre-trained supernet and fine-tuning for a few epochs, enabling a comparable landscape exploration to what we have presented for DARTS.

\subsection{A$^2$M Algorithm Evaluation}

To demonstrate the generalization improvements introduced by A$^2$M, we evaluate its impact on seven state-of-the-art DARTS-based methods: DARTS ($1^{st}$ order)~\cite{liu_darts_2019}, $\beta$-DARTS~\cite{betadarts}, $\Lambda$-DARTS~\cite{lambdadarts}, DARTS-PT\footnote{For A$^2$-DARTS-PT, we apply A$^2$M both during the architecture search phase and the final model selection phase, where a tuning of the supernet is performed.}~\cite{wang2021rethinking}, DARTS-~\cite{darts-}, SDARTS-RS~\cite{sdarts-adv}, and PC-DARTS~\cite{pc-darts}.
We evaluate these methods and their A$^2$-enhanced variants on both NAS-Bench-201 and DARTS search spaces across three benchmark datasets: CIFAR-10, CIFAR-100, and ImageNet-16-12.
All the evaluations are performed on an NVIDIA A40 GPU.
We highlight that the architecture search is conducted only on CIFAR-10, and the discovered genotype is subsequently evaluated on the other datasets.
To determine the optimal~$\rho_{\alpha}$ value for A$^2$M, we perform a search on CIFAR-10 only,
reported in Appendix~\ref{appendix:rho}.
A$^2$M introduces only a modest increase in search cost, ranging between~$1.1 \times $  and~$1.3 \times $  times the one of the original methods. For each experiment, we computed the results for 5 different seeds and reported the average value and the standard deviation. 
\begin{table*}[h!]
    \caption{Comparison of test accuracy on NAS-Bench-201 (upper table) and DARTS search space (lower table) across CIFAR-10, CIFAR-100, and ImageNet-16-120 datasets for multiple SotA methods, showing standard and A$^2$-enhanced results. The results are averaged over 5 different seeds. *DARTS and A$^2$-DARTS have been run without data augmentation on the validation set on NAS-Bench-201. The best result for each method between the standard and the A$^2$-enhanced version on a dataset is reported in \textbf{bold}. The best result for each search space on a dataset is \underline{underlined}. \textit{Optimal} refers to the best NN of NAS-Bench-201. \textit{Avg Test Acc} refers to the average test accuracy by averaging over all the reported methods and search space for each dataset. \textit{Avg Gain} refers to the gain in \textit{Avg Test Acc} for each dataset. }
    \label{tab:test}
\vskip 0.15in
\begin{center}
\begin{small}
\begin{sc}    
\scalebox{0.85}{
    \begin{tabular}{ccccccc}
    \toprule
    \multirow{2}{*}{\textbf{Model}} & \multicolumn{2}{c}{\textbf{CIFAR-10 Test Acc (\%)}} & \multicolumn{2}{c}{\textbf{CIFAR-100 Test Acc (\%)}} & \multicolumn{2}{c}{\textbf{ImageNet-16-120 Test Acc (\%)}} \\
    \cline{2-7}
     & \textbf{Standard} & \textbf{A$^2$-} & \textbf{Standard} & \textbf{A$^2$-} & \textbf{Standard} & \textbf{A$^2$-} \\
    \midrule
    DARTS ($1^{st}$)*~\cite{liu_darts_2019} & 72.85$\pm$35.21 & \textbf{92.39$\pm$0.52} & 49.44$\pm$27.39 & \textbf{68.79$\pm$1.61} & 26.24$\pm$15.33 & \textbf{41.88$\pm$2.21} \\
    $\beta$-DARTS~\cite{betadarts} & \underline{\textbf{94.15$\pm$0.29}} & \underline{\textbf{94.15$\pm$0.29}} & \underline{\textbf{72.77$\pm$0.83}} & \underline{\textbf{72.77$\pm$0.83}} & \underline{\textbf{45.66$\pm$0.92}} & \underline{\textbf{45.66$\pm$0.92}} \\
    $\Lambda$-DARTS~\cite{lambdadarts} & 92.69$\pm$1.90 & \textbf{93.96$\pm$0.47} & 69.99$\pm$3.78 & \textbf{72.05$\pm$1.73} & 43.00$\pm$3.36 & \textbf{45.27$\pm$1.27} \\
    DARTS-PT\cite{wang2021rethinking}& 88.40$\pm$0.0& \textbf{92.31$\pm$0.0} & 61.35$\pm$0.0 & \textbf{68.19$\pm$0.0} & 34.62$\pm$0.0 &  \textbf{41.24$\pm$0.0} \\
    DARTS-~\cite{darts-} & \textbf{93.76$\pm$0.43} & \textbf{93.76$\pm$0.43} & \textbf{71.11$\pm$1.60} & \textbf{71.11$\pm$1.60} & \textbf{41.44$\pm$0.90} & \textbf{41.44$\pm$0.90} \\
    SDARTS-RS~\cite{chen20} & 80.57$\pm$0.30& \textbf{84.33$\pm$0.18} & 47.93$\pm$0.43 & \textbf{54.96$\pm$0.33} & \textbf{26.29$\pm$0.72} &  25.62$\pm$0.26 \\
    PC-DARTS~\cite{pc-darts} & 70.92$\pm$0.35 & \textbf{91.95$\pm$0.16} & 38.97$\pm$0.80 & \textbf{66.47$\pm$0.40} & 18.41$\pm$0.70 &  \textbf{39.59$\pm$0.97} \\
    Optimal~\cite{nasbench201} & \multicolumn{2}{c}{94.37} & \multicolumn{2}{c}{73.51} & \multicolumn{2}{c}{46.71} \\
    \midrule
    \midrule
    DARTS ($1^{st}$)~\cite{liu_darts_2019} & 97.00$\pm$0.14 & \textbf{97.20$\pm$0.15} & 82.30$\pm$0.30 & \textbf{82.50$\pm$0.79} 
    & 53.06$\pm$0.53 & \textbf{55.03$\pm$0.37} \\
    $\beta$-DARTS~\cite{betadarts} & 96.83$\pm$0.15 & \textbf{96.94$\pm$0.16} & 81.85$\pm$0.68 & \textbf{82.48$\pm$0.27}
    & 53.92$\pm$0.47 & \textbf{54.45$\pm$0.42} \\
    $\Lambda$-DARTS~\cite{lambdadarts} & 97.05$\pm$0.39 & \textbf{97.14$\pm$0.32} & 82.92$\pm$0.32 & \textbf{83.16$\pm$0.42} 
    & 53.62$\pm$0.48 & \underline{\textbf{58.92$\pm$0.33}} \\
    DARTS-PT~\cite{wang2021rethinking} & 97.36$\pm$0.09 & \underline{\textbf{97.42$\pm$0.07}} & 83.53$\pm$0.23 & \underline{\textbf{83.60$\pm$0.21}}
    & 54.00$\pm$0.39 & \textbf{54.28$\pm$0.35} \\
    DARTS-~\cite{darts-}& 97.30$\pm$0.15  & \textbf{97.36$\pm$0.12} & \textbf{82.43$\pm$0.31} & 82.32$\pm$0.24 
    & 53.78$\pm$0.51 & \textbf{54.32$\pm$0.43} \\
    SDARTS-RS~\cite{sdarts-adv}& 97.33$\pm$0.03 & \textbf{97.38$\pm$0.04} & 82.50$\pm$0.22 & \textbf{82.67$\pm$0.24}
    & 55.48$\pm$0.30 & \textbf{55.70$\pm$0.38} \\
    PC-DARTS~\cite{pc-darts}& 97.15$\pm$0.18 & \textbf{97.20$\pm$0.13} & 82.18$\pm$0.27 & \textbf{82.60$\pm$0.25}
    & \textbf{57.38$\pm$0.46} & 54.33$\pm$0.54 \\
     \midrule
    \midrule
    Avg Test Acc (\%) & 90.95 & \textbf{94.54} & 70.66 & \textbf{75.26} 
    & 44.06 & \textbf{47.70} \\
    \midrule
    Avg Gain (\%)  & \multicolumn{2}{c}{+3.60} & \multicolumn{2}{c}{+4.60} & \multicolumn{2}{c}{+3.64} \\
    \bottomrule
    \bottomrule
    \end{tabular}
    }
\end{sc}
\end{small}
\end{center}
\vskip -0.1in
\end{table*}

\subsection{Results on NAS-Bench-201 Search Space}

For experiments in the NAS-Bench-201 search space, we follow the original search settings of DARTS on NAS-Bench-201 \cite{nasbench201}.
Additionally, for~$\Lambda$-DARTS, we limit the search phase to 50 epochs to ensure consistency with the search runs of other NAS methods~\cite{lambdadarts}.
The results are presented in the first section of Table~\ref{tab:test}, where we compare test accuracies between the original methods (labeled as Standard) and their A$^2$-enhanced variants for each dataset. As in prior DARTS-based studies, validation accuracies are reported separately in Appendix~\ref{appendix:additional}.
To provide an upper bound on achievable performance within the search space, we include the optimal architecture in NAS-Bench-201 as a baseline, labeled OPTIMAL. 
As shown in the upper part of Table~\ref{tab:test}, A$^2$M enhances performance across all evaluated methods apart from $\beta$-DARTS and DARTS-, where performance remains unchanged. 
We observed that on NAS-Bench-201, DARTS and A$^2$-DARTS consistently yield improved performance when data augmentation is disabled on the validation set, as discussed in Appendix~\ref{appendix:dataaug}. We report the results of both algorithms with this setting.
In particular, results show a huge improvement in DARTS and PC-DARTS, with a maximum of a more than double (2.15$\times$) increase in accuracy of PC-DARTS on ImageNet-16-120.
Additionally, we highlight that $\beta$-DARTS and $A^2$-$\beta$-DARTS find the same model for each run of the five seeds evaluated. The same applies to DARTS- and $A^2$-DARTS-. We report all hyper-parameters for experimental reproducibility in Appendix~\ref{sec:nasbench201_params}, with the search for the A$^2$M hyperparameter $\rho_{\alpha}$ reported in Appendix~\ref{appendix:rho}.

The overall success of A$^2$M-enhanced methods in improving generalization, as detailed in Table~\ref{tab:test} for the NAS-Bench-201 space, aligns with our broader geometric findings. Our path analysis (Figure~\ref{fig:paths}) demonstrates that high-performing architectures in NAS-Bench-201 generally reside in flatter regions with lower accuracy barriers. While some A$^2$M-enhanced methods converged to the same robust architectures as their standard counterparts (e.g., $\beta$-DARTS, DARTS--), the significant improvements seen with other methods (e.g., DARTS, PC-DARTS) when enhanced by A$^2$M support the conclusion that explicitly targeting these geometrically favorable flat regions is a beneficial strategy for discovering better generalizing minima.

\subsection{Results on DARTS Search Space}

For experiments conducted in the DARTS search space, we follow the original search settings and the evaluation settings for all discovered networks of DARTS for CIFAR-10~\cite{liu_darts_2019}. The hyperparameters of the search and the training are shown in Table \ref{tab:settingsearchdarts} and Table \ref{tab:settingtraindarts}, respectively.
Moreover, the architecture search is performed on CIFAR-10, and the resulting genotype is then evaluated on various datasets by training the corresponding neural networks from scratch. we highlight that during the search phase, the explored networks are smaller in terms of size with respect to the final trained network to balance efficiency and performance of the NAS. This is shown by the different number of channels and cells stacked in Table \ref{tab:settingsearchdarts} and Table \ref{tab:settingtraindarts}.
The comparison results are summarized in the lower part of Table~\ref{tab:test}, which reports the test accuracies for both the original methods (labeled as Standard) and their A$^2$-enhanced variants across all datasets. For $\beta$-DARTS, we reproduced the procedure based on its original GitHub repository to the best of our efforts, with detailed reproduction steps outlined in Appendix~\ref{betaweights}.


The results in the lower part of Table~\ref{tab:test} demonstrate that A$^2$M improves performance across all evaluated methods on all datasets, with the exception of DARTS- on CIFAR-100 and PC-DARTS on ImageNet-16-120.
We highlight that the \textit{best methods in terms of test accuracy of the final model are all A$^2$-enhanced ones}: A$^2$-DARTS-PT on CIFAR-10 and CIFAR-100 and A$^2$-$\Lambda$-DARTS on ImageNet-16-120. We report all hyperparameters for experimental reproducibility in Appendix~\ref{sec:darts_params}, with the search for the A$^2$M hyperparameter $\rho_{\alpha}$ reported in Appendix~\ref{appendix:rho}.

We observe a clear correlation: architectures found by A$^2$M, which are designed to lie in flatter regions (as evidenced by their lower accuracy barriers in Figure~\ref{fig:pathsDARTS} for the DARTS space), consistently achieve higher test accuracies as shown in Table~\ref{tab:test}. This supports the conclusion that optimizing for flatness in architecture space, as A$^2$M does, indeed leads to discovering better generalizing minima, validating the practical benefit of our geometric insights.

\section{Conclusions}
\label{sec:conclusions}

By conducting an extensive and quantitative analysis of the geometric properties of architecture spaces used in differentiable NAS methods, such as NAS-Bench-201 and DARTS, we establish clear parallels between architecture landscapes and loss landscapes in weight space. Notably, by studying architecture neighborhoods and introducing a principled way of computing paths in the architecture landscape, we show that both search spaces exhibit locality and flatness properties, with flat regions corresponding to clusters of high-performing architectures, while architectures with lower performances are separated by higher accuracy barriers, and therefore more isolated. Our results shed light on the relationship between the geometrical properties of the architecture space and network performance, and yield practical implications for improving the effectiveness of differentiable NAS optimization algorithms. 

A$^2$M, our proposed algorithmic framework introducing a new architecture gradient update step in the DARTS bi-level optimization problem, can be easily applied to DARTS-based methods, and shows substantial performance gains obtained on several benchmark datasets, several DARTS-based methods and search spaces, demonstrating the practical benefits of explicitly targeting flat regions in the architecture space.
In fact, by explicitly targeting flatness, A$^2$M consistently outperforms state-of-the-art DARTS-based algorithms, leading to improved generalization and variance reduction. 

However, this work also opens several avenues for future research and highlights areas for further improvement.
The \textit{scope of our geometric analysis and neighbor definitions} can be expanded. Our current neighborhood definition (Section~\ref{subsec:neighborhood}) primarily considers modifications within a fixed cell structure. Future work should explore extending this to include macro-structural changes, such as the dynamic addition or removal of nodes, thereby altering the dimensionality of architectural encodings and capturing a broader range of structural variations. This would provide a more comprehensive understanding of the architecture landscape.
Our \textit{visualization and landscape analysis methodologies} have proven insightful for NAS-Bench-201 and DARTS. Applying these techniques to a wider array of NAS benchmarks, including those for different tasks (e.g., NLP via NAS-Bench-NLP~\cite{NAS-Bench-NLP}) or those considering other objectives like adversarial robustness (e.g., NAS-RobBench-201~\cite{RobustNAS}), will be crucial for validating the generality of the observed geometric properties and the benefits of targeting flatness.

A key direction will be to \textit{address the computational overhead} introduced by the A$^2$M gradient update. While currently modest, optimizing this aspect, potentially by incorporating strategies from more advanced Sharpness-Aware Minimization variants (e.g., ASAM~\cite{ASAM}, GA-SAM~\cite{ga-sam}) that offer adaptive perturbation or more efficient gradient estimation, could enhance A$^2$M scalability, especially for extremely large search spaces or prolonged search phases. The \textit{A$^2$M algorithm itself}, in its present form, is designed for differentiable NAS (D-NAS) frameworks due to its reliance on continuous architecture parameters and gradient-based updates. While it can be readily integrated into other D-NAS methods, a significant and exciting challenge lies in \textit{extending the principles of flatness-aware search to discrete NAS paradigms}. This would necessitate the development of novel algorithms, distinct from the current A$^2$M gradient formulation, that can effectively bias discrete search strategies (e.g., evolutionary algorithms, reinforcement learning, or supernet path sampling in OFA-like models~\cite{ofa}) towards flatter, more robust regions of the discrete architecture space.
Finally, further \textit{refinement of the A$^2$M mechanism} could involve integrating more sophisticated SAM-like logic, such as dynamically adapting the perturbation radius ($\rho_{\alpha}$) based on local landscape characteristics during the NAS process, rather than using a fixed value. Investigating the interplay between flatness in architecture space and other desirable properties, such as Out-Of-Distribution or Adversarial robustness, or adaptability to concept drift, also remains a fertile ground for future exploration, potentially leading to novel and improved NAS algorithms.

\section*{Acknowledgements}
This paper is supported by the PNRR-PE-AI FAIR project funded by the NextGeneration EU program.

\bibliography{anonymous-submission}
\bibliographystyle{ieeetr}

\newpage
\appendix
\onecolumn

\section{Warm up: DARTS bi-level optimization and USAM update}

In this section, we provide additional background on the DARTS bi-level optimization problem and on the USAM update, that will both be instrumental to analytically derive our newly proposed A$^2$M architecture gradient update, introduced in Eq.~\eqref{eq:A$^2$M}.

\subsection{DARTS Approximated Gradient}

By defining the one-weight-step approximated gradient: 
\begin{equation}
w'\left(\alpha\right)  = w-\xi\nabla_{w}\mathcal{L}_{\text{train}}\left(w,\alpha\right)\label{eq:wprime}
\end{equation}
and the chain rule
\begin{equation}
\nabla_{\alpha}\mathcal{L}\left(w'\left(\alpha\right)\right)=\nabla_{\alpha}w'\left(\alpha\right)\nabla_{w'}\mathcal{L}\left(w'\right)\label{eq:chainrule_def}
\end{equation}
we can write the one-weight-step approximated DARTS gradient as:
\begin{align}
\nabla_{\alpha}\mathcal{L}_{\text{val}}\left(w^{*}\left(\alpha\right),\alpha\right) & \approx  \nabla_{\alpha}\mathcal{L}_{\text{val}}\left(w'\left(\alpha\right),\alpha\right)\\
 & =  \nabla_{\alpha}\mathcal{L}_{\text{val}}\left(w',\alpha\right)+\nabla_{\alpha}w'\left(\alpha\right)\nabla_{w'}\mathcal{L}_{\text{val}}\left(w',\alpha\right)\label{eq:chainrule}\\
 & =  \nabla_{\alpha}\mathcal{L}_{\text{val}}\left(w',\alpha\right)+\nabla_{\alpha}\left(w-\xi\nabla_{w}\mathcal{L}_{\text{train}}\left(w,\alpha\right)\right)\nabla_{w'}\mathcal{L}_{\text{val}}\left(w',\alpha\right)\\
 & =  \nabla_{\alpha}\mathcal{L}_{\text{val}}\left(w',\alpha\right)-\xi\nabla_{\alpha,w}^{2}\mathcal{L}_{\text{train}}\left(w,\alpha\right)\nabla_{w'}\mathcal{L}_{\text{val}}\left(w',\alpha\right)\\
 & \approx  \nabla_{\alpha}\mathcal{L}_{\text{val}}\left(w',\alpha\right)-\xi\left(\frac{\nabla_{\alpha}\mathcal{L}_{\text{train}}\left(w^{+},\alpha\right)-\nabla_{\alpha}\mathcal{L}_{\text{train}}\left(w^{-},\alpha\right)}{2\epsilon}\right)\label{eq:finitediff}
\end{align}
where in Eq.~(\ref{eq:chainrule}) we have summed up the contributions
to the gradient coming from the first and second argument of $\mathcal{L}_{val}$.
for the derivative with respect to the second argument we have used the chain
rule Eq.(\ref{eq:chainrule_def}), and in Eq.(\ref{eq:finitediff})
we have used the finite difference approximation
\begin{equation}
\nabla_{\alpha,w}^{2}\mathcal{L}_{train}\left(w,\alpha\right)\approx\frac{\nabla_{\alpha}\mathcal{L}_{train}\left(w^{+},\alpha\right)-\nabla_{\alpha}\mathcal{L}_{train}\left(w^{-},\alpha\right)}{2\epsilon\nabla_{w'}\mathcal{L}_{val}\left(w',\alpha\right)}
\end{equation}
with
\begin{equation}
w^{\pm}=w\pm\epsilon\nabla_{w'}\mathcal{L}_{val}\left(w',\alpha\right).
\end{equation}

\subsection{The USAM Update}

We recall the USAM~\cite{usam} gradient update. In USAM, the SAM gradient step is simplified by avoiding normalization, reducing computational complexity, and has been empirically shown to perform well~\cite{usam}.
Eq.~\eqref{eq:usam} can be rewritten as:
\begin{equation}
\nabla f\left(x_{k}\right)\stackrel{USAM}{\longrightarrow}\nabla f\left(x_{k}+\rho\nabla f(x_{k})\right)
\end{equation}
or
\begin{equation}
\nabla f\left(x_{k}\right)\stackrel{USAM}{\longrightarrow}\nabla f\left(\tilde{x}_{k}\left(x_{k}\right)\right)\quad\text{with}\quad\tilde{x}_{k}\left(x_{k}\right)=x_{k}+\rho\nabla f(x_{k})
\end{equation}

\section{Derivation of the A$^2$M Gradient Update}
\label{appendix:A$^2$M}

In this section, we derive our newly proposed A$^2$M gradient update step, introduced in Eq.~\eqref{eq:A$^2$M}.
As a preliminary step, let us write the USAM update specifically for $\alpha$:
\begin{equation}
\nabla_{\alpha}\mathcal{L}_{val}\left(w'\left(\alpha\right),\alpha\right)\stackrel{USAM}{\longrightarrow}\nabla_{\alpha}\mathcal{L}_{val}\left(w'\left(\tilde{\alpha}\left(\alpha\right)\right),\tilde{\alpha}\left(\alpha\right)\right)
\end{equation}
where 
\begin{align}
\tilde{\alpha}\left(\alpha\right) & =  \alpha+\rho_{\alpha}\nabla_{\alpha}\mathcal{L}_{val}\left(w'\left(\alpha\right),\alpha\right)\\
 & =  \alpha+\rho_{\alpha}\nabla_{\alpha}\mathcal{L}_{val}\left(w',\alpha\right)-\rho_{\alpha}\xi\left(\frac{\nabla_{\alpha}\mathcal{L}_{train}\left(w^{+},\alpha\right)-\nabla_{\alpha}\mathcal{L}_{train}\left(w^{-},\alpha\right)}{2\epsilon}\right)\label{eq:alphatilde}
\end{align}
where in Eq.~(\ref{eq:alphatilde}) we have used Eq.~(\ref{eq:finitediff}).

\subsection{Gradient With Respect To the First Argument}

In order to obtain the total gradient, we have to differentiate $\mathcal{L}_{val}\left(\cdot,\cdot\right)$
separately with respect to its first and second argument and sum the two gradients.
Let us calculate in this section the gradient with respect to the first argument,
which involves the definition of $w'\left(\alpha\right)$ Eq.~\eqref{eq:wprime}:
\begin{align}
\nabla_{\alpha}\mathcal{L}_{val}\left(w'\left(\tilde{\alpha}\left(\alpha\right)\right),\tilde{\alpha}\left(\alpha\right)\right)\Big|_{1^{\text{st}}\text{ argument}} & =  \left(\nabla_{\alpha}w'\left(\tilde{\alpha}\left(\alpha\right)\right)\right)\nabla_{w'}\mathcal{L}_{val}\left(w',\tilde{\alpha}\right)\\
 & =  \left(\nabla_{\alpha}\tilde{\alpha}\left(\alpha\right)\right)\nabla_{\tilde{\alpha}}w'\left(\tilde{\alpha}\right)\nabla_{w'}\mathcal{L}_{val}\left(w',\tilde{\alpha}\right)
\end{align}
where
\begin{align}
\nabla_{\alpha}\tilde{\alpha}\left(\alpha\right) & =  \nabla_{\alpha}\left(\alpha+\rho_{\alpha}\nabla_{\alpha}\mathcal{L}_{val}\left(w',\alpha\right)-\rho_{\alpha}\xi\left(\frac{\nabla_{\alpha}\mathcal{L}_{train}\left(w^{+},\alpha\right)-\nabla_{\alpha}\mathcal{L}_{train}\left(w^{-},\alpha\right)}{2\epsilon}\right)\right)\\
 & =  1+\rho_{\alpha}\nabla_{\alpha}^{2}\mathcal{L}_{val}\left(w',\alpha\right)-\rho_{\alpha}\xi\left(\frac{\nabla_{\alpha}^{2}\mathcal{L}_{train}\left(w^{+},\alpha\right)-\nabla_{\alpha}^{2}\mathcal{L}_{train}\left(w^{-},\alpha\right)}{2\epsilon}\right)\\
\end{align}
and
\begin{align}
\nabla_{\tilde{\alpha}}w'\left(\tilde{\alpha}\right) & =  \nabla_{\tilde{\alpha}}\left(w-\xi\nabla_{w}\mathcal{L}_{train}\left(w,\tilde{\alpha}\right)\right)\\
 & =  \xi\nabla_{w,\tilde{\alpha}}^{2}\mathcal{L}_{train}\left(w,\tilde{\alpha}\right)
\end{align}
so that we obtain
\begin{equation}
\begin{aligned}
\nabla_{\alpha}\mathcal{L}_{val}&\left(w'\left(\tilde{\alpha}\left(\alpha\right)\right),\tilde{\alpha}\left(\alpha\right)\right)\Big|_{1^{\text{st}}\text{ argument}}  =  \Bigg(1+\rho_{\alpha}\nabla_{\alpha}^{2}\mathcal{L}_{val}\left(w',\alpha\right)\\
&-\rho_{\alpha}\xi\left(\frac{\nabla_{\alpha}^{2}\mathcal{L}_{train}\left(w^{+},\alpha\right)-\nabla_{\alpha}^{2}\mathcal{L}_{train}\left(w^{-},\alpha\right)}{2\epsilon}\right)\Bigg)
 \xi\nabla_{w,\tilde{\alpha}}^{2}\mathcal{L}_{train}\left(w,\tilde{\alpha}\right)\nabla_{w'}\mathcal{L}_{val}\left(w',\tilde{\alpha}\right).
\end{aligned}
\end{equation}
Therefore, as we uncover that this gradient is linear in the weight learning rate $\xi$, in the approximation in which it is set to~$\xi = 0$, i.e. a DARTS-like first order approximation, the entire gradient of~$\mathcal{L}_{val}$ with respect to the first argument is set to zero and does not contribute to the A$^2$M gradient.

\subsection{Gradient With Respect To the Second Argument}

Let us compute the gradient with respect to the second argument. We apply the chain rule to the second argument of~$\mathcal{L}_{val}\left(\cdot,\cdot\right)$:
\begin{equation}
\nabla_{\alpha}\mathcal{L}_{val}\left(w'\left(\tilde{\alpha}\left(\alpha\right)\right),\tilde{\alpha}\left(\alpha\right)\right)\Big|_{2^{\text{nd}}\text{ argument}}=\left(\nabla_{\alpha}\tilde{\alpha}\left(\alpha\right)\right)\nabla_{\tilde{\alpha}}\mathcal{L}_{val}\left(w'\left(\tilde{\alpha}\right),\tilde{\alpha}\right)
\end{equation}
such that
\begin{align}
&\nabla_{\alpha}\mathcal{L}_{val}\left(w'\left(\tilde{\alpha}\left(\alpha\right)\right),\tilde{\alpha}\left(\alpha\right)\right)=\label{eq:alpha_gradient}\\
 & = \left(\nabla_{\alpha}\tilde{\alpha}\left(\alpha\right)\right)\nabla_{\tilde{\alpha}}\mathcal{L}_{val}\left(w'\left(\tilde{\alpha}\right),\tilde{\alpha}\right)\nonumber \\
 & =\left(1+\rho_{\alpha}\nabla_{\alpha}^{2}\mathcal{L}_{val}\left(w',\alpha\right)-\rho_{\alpha}\xi\left(\frac{\nabla_{\alpha}^{2}\mathcal{L}_{train}\left(w^{+},\alpha\right)-\nabla_{\alpha}^{2}\mathcal{L}_{train}\left(w^{-},\alpha\right)}{2\epsilon}\right)\right)\nabla_{\tilde{\alpha}}\mathcal{L}_{val}\left(w'\left(\tilde{\alpha}\right),\tilde{\alpha}\right)\nonumber \\
 & =\nabla_{\tilde{\alpha}}\mathcal{L}_{val}\left(w'\left(\tilde{\alpha}\right),\tilde{\alpha}\right)\\
 &\quad+\left(\rho_{\alpha}\nabla_{\alpha}^{2}\mathcal{L}_{val}\left(w',\alpha\right)-\rho_{\alpha}\xi\left(\frac{\nabla_{\alpha}^{2}\mathcal{L}_{train}\left(w^{+},\alpha\right)-\nabla_{\alpha}^{2}\mathcal{L}_{train}\left(w^{-},\alpha\right)}{2\epsilon}\right)\right)\nabla_{\tilde{\alpha}}\mathcal{L}_{val}\left(w'\left(\tilde{\alpha}\right),\tilde{\alpha}\right)\nonumber 
\end{align}

Now we can apply the finite difference approximation to the terms
\begin{align}
\nabla_{\alpha}^{2}\mathcal{L}_{val}\left(w',\alpha\right) & \approx  \frac{\nabla_{\alpha}\mathcal{L}_{val}\left(w',\alpha_{1}^{+}\right)-\nabla_{\alpha}\mathcal{L}_{val}\left(w',\alpha_{1}^{-}\right)}{2\epsilon_{1}\nabla_{\tilde{\alpha}}\mathcal{L}_{val}\left(w'\left(\tilde{\alpha}\right),\tilde{\alpha}\right)}\\
\nabla_{\alpha}^{2}\mathcal{L}_{train}\left(w^{+},\alpha\right) & \approx  \frac{\nabla_{\alpha}\mathcal{L}_{train}\left(w^{+},\alpha_{2}^{+}\right)-\nabla_{\alpha}\mathcal{L}_{train}\left(w^{+},\alpha_{2}^{-}\right)}{2\epsilon_{2}\nabla_{\tilde{\alpha}}\mathcal{L}_{val}\left(w'\left(\tilde{\alpha}\right),\tilde{\alpha}\right)}\\
\nabla_{\alpha}^{2}\mathcal{L}_{train}\left(w^{-},\alpha\right) & \approx  \frac{\nabla_{\alpha}\mathcal{L}_{train}\left(w^{-},\alpha_{3}^{+}\right)-\nabla_{\alpha}\mathcal{L}_{train}\left(w^{-},\alpha_{3}^{-}\right)}{2\epsilon_{3}\nabla_{\tilde{\alpha}}\mathcal{L}_{val}\left(w'\left(\tilde{\alpha}\right),\tilde{\alpha}\right)}
\end{align}
where
\begin{equation}
\alpha_{i}^{\pm}=\alpha\pm\epsilon_{i}\nabla_{\tilde{\alpha}}\mathcal{L}_{val}\left(w'\left(\tilde{\alpha}\right),\tilde{\alpha}\right)\quad\text{for}\quad i={1,2,3}
\end{equation}
such that finally we obtain that the gradient Eq.~\eqref{eq:alpha_gradient} is:
\begin{align}
&\nabla_{\alpha}\mathcal{L}_{val}\left(w'\left(\tilde{\alpha}\left(\alpha\right)\right),\tilde{\alpha}\left(\alpha\right)\right)= \label{eq:final1}\\
& \nabla_{\tilde{\alpha}}\mathcal{L}_{val}\left(w'\left(\tilde{\alpha}\right),\tilde{\alpha}\right)
+\rho_{\alpha}\frac{\nabla_{\alpha}\mathcal{L}_{val}\left(w',\alpha^{+}\right)-\nabla_{\alpha}\mathcal{L}_{val}\left(w',\alpha^{-}\right)}{2\epsilon}+\\
&-\rho_{\alpha}\xi\left(\frac{1}{2\epsilon}\left(\frac{\nabla_{\alpha}\mathcal{L}_{train}\left(w^{+},\alpha^{+}\right)-\nabla_{\alpha}\mathcal{L}_{train}\left(w^{+},\alpha^{-}\right)}{2\epsilon}-\frac{\nabla_{\alpha}\mathcal{L}_{train}\left(w^{-},\alpha^{+}\right)-\nabla_{\alpha}\mathcal{L}_{train}\left(w^{-},\alpha^{-}\right)}{2\epsilon}\right)\right).
\end{align}

We can obtain the DARTS-like first-order approximation by setting
the weight learning rate $\xi$ to zero and we obtain:
\begin{align}
\nabla_{\alpha}\mathcal{L}_{\text{val}}&\left(w'\left(\tilde{\alpha}\left(\alpha\right)\right),\tilde{\alpha}\left(\alpha\right)\right)  =  \nabla_{\alpha}\mathcal{L}_{\text{val}}\left(w,\tilde{\alpha}_{\xi=0}\left(\alpha\right)\right)\\
  = & \nabla_{\tilde{\alpha}_{\xi=0}}\mathcal{L}_{\text{val}}\left(w,\tilde{\alpha}_{\xi=0}\right)+\rho_{\alpha}\frac{\nabla_{\alpha}\mathcal{L}_{\text{val}}\left(w,\alpha_{\xi=0}^{+}\right)-\nabla_{\alpha}\mathcal{L}_{\text{val}}\left(w,\alpha_{\xi=0}^{-}\right)}{2\epsilon}\\
  = & 
\nabla_{\tilde{\alpha}_{\xi=0}}\mathcal{L}_{\text{val}}\left(w,\tilde{\alpha}_{\xi=0}\right)
+\rho_{\alpha}\frac{\nabla_{\alpha_{\xi=0}^{+}}\mathcal{L}_{\text{val}}\left(w,\alpha_{\xi=0}^{+}\right)-\nabla_{\alpha_{\xi=0}^{-}}\mathcal{L}_{\text{val}}\left(w,\alpha_{\xi=0}^{-}\right)}{2\epsilon}
\left(1+\mathcal{O}\left(\epsilon\right)\right)
\end{align}
with
\begin{align}
\alpha_{\xi=0}^{\pm} & =  \alpha\pm\epsilon\nabla_{\tilde{\alpha}_{\xi=0}}\mathcal{L}_{\text{val}}\left(w,\tilde{\alpha}_{\xi=0}\right)\\
\tilde{\alpha}_{\xi=0} & =  \alpha+\rho_{\alpha}\nabla_{\alpha}\mathcal{L}_{\text{val}}\left(w,\alpha\right)
\end{align}
which is the A$^2$M formula reported in the main text, Eqs.~\eqref{eq:A$^2$M}.

By applying the perturbation only to~$\alpha$, we specifically target flatness in the architecture parameter space without increasing the computational burden significantly because the number of architecture parameters is significantly smaller than the number of network weights.
The choice of $\rho_{\alpha}$ is critical; it controls the extent of the neighborhood considered for sharpness minimization. We select $\rho_{\alpha}$ based on validation performance, as shown in Appendix~\ref{appendix:rho}. 
A$^2$M is compatible with existing DARTS implementations and can be integrated with other NAS methods based on a continuous relaxation of the architecture space.

\section{Experimental Reproducibility and Hyperparameter Settings}
\label{app:hyperparams}

This appendix provides a comprehensive list of the key hyperparameters used in our experiments to ensure full reproducibility of our results. The following subsections detail the parameters for experiments conducted on the NAS-Bench-201 search space, the DARTS search space, and the specific values selected for our proposed $\rho_{\alpha}$ hyperparameter.
We present further discussion on data augmentation within NAS-Bench-201 and outline our implementation of \texorpdfstring{$\beta$}{beta}-DARTS regularization for the DARTS search space.

\subsection{Hyperparameters for NAS-Bench-201 Experiments}
\label{sec:nasbench201_params}

This subsection details the hyperparameters employed for the search phase on the NAS-Bench-201 search space across the CIFAR-10, CIFAR-100, and ImageNet-16-120 datasets. These parameters are summarized in Table~\ref{tab:nasbench201_paramss}.
\begin{table}[h]
\centering
\caption{Hyperparameters for Search Phase on CIFAR-10 on NAS-Bench-201 search space}
\scalebox{0.8}{
\begin{tabular}{|l|l|}
\hline
\textbf{Hyperparameter} & \textbf{Value / Setting} \\
\hline
Dataset & CIFAR-10 (split into train/val) \\
Epochs & 50 \\
Batch size & 64 \\
Optimizer (weights) & SGD (momentum = 0.9, weight decay = $5 \times 10^{-4}$) \\
Optimizer (arch parameters) & Adam (lr = $3 \times 10^{-4}$, betas = (0.5, 0.999), weight decay = $10^{-3}$) \\
Learning rate & decay from 0.025 to 0.001 (cosine annealing) \\
Gradient clipping & 5.0 \\
Initial channels & 16 \\
Number of cells & 5 \\
Weight decay (arch params) & $10^{-3}$ \\
Weight decay (model weights) & $5 \times 10^{-4}$ \\
Data augmentation & Random Horizontal Flipping, Random Cropping with Padding, and Normalization \\
$\rho_\alpha$ (only for $A^2M$-enhanced methods) &  $10^{-2}$ \\
\hline
\end{tabular}
}
\label{tab:nasbench201_paramss}
\end{table}

\subsection{Hyperparameters for DARTS Experiments}
\label{sec:darts_params}

In Table~\ref{tab:settingsearchdarts} we outline the hyperparameters used for the architectural search and in Table~\ref{tab:settingtraindarts} the subsequent full training evaluation stages within the DARTS search space on the CIFAR-10, CIFAR-100, and ImageNet-16-120 datasets.
\begin{table}[h]
\centering
\caption{Hyperparameters for Search Phase on CIFAR-10 on DARTS search space}
\scalebox{0.8}{
\begin{tabular}{|l|l|}
\hline
\textbf{Hyperparameter} & \textbf{Value / Setting} \\
\hline
Dataset & CIFAR-10 (split into train/val) \\
Epochs & 50 \\
Batch size & 64 \\
Optimizer (weights) & SGD (momentum = 0.9, weight decay = $3 \times 10^{-4}$) \\
Optimizer (arch parameters) & Adam (lr = $3 \times 10^{-4}$, betas = (0.5, 0.999), weight decay = $10^{-3}$) \\
Learning rate & decay from 0.025 to 0.001 (cosine annealing) \\
Gradient clipping & 5.0 \\
Initial channels & 16 \\
Number of cells & 8 \\
Weight decay (arch parameters) & $10^{-3}$ \\
Weight decay (model weights) & $3 \times 10^{-4}$ \\
Data augmentation & Random Horizontal Flipping, Random Cropping with Padding, and Normalization \\
$\rho_\alpha$ (only for $A^2M$-enhanced methods) &  $10^{-1}$ \\
\hline
\end{tabular}
}
\label{tab:settingsearchdarts}
\end{table}

\begin{table}[H]
\centering
\caption{Hyperparameters for Final Training Phase on a dataset on DARTS search space}
\scalebox{0.9}{
\begin{tabular}{|l|l|}
\hline
\textbf{Hyperparameter} & \textbf{Value / Setting} \\
\hline
Dataset & Full training set of the dataset \\
Epochs & 600 \\
Batch size & 96 \\
Optimizer & SGD (momentum = 0.9, weight decay = $3 \times 10^{-4}$) \\
Learning rate & decay from 0.025 to 0.001 (cosine annealing) \\
Gradient clipping & 5.0 \\
Initial channels & 36 \\
Number of cells & 20 \\
Auxiliary towers & True (weight = 0.4) \\
Drop path probability & 0.2 (linearly increased) \\
Cutout & True (length = 16) \\
Weight decay & $3 \times 10^{-4}$ \\
Data augmentation & Random Horizontal Flipping, Random Cropping with Padding, and Normalization \\
\hline
\end{tabular}
}
\label{tab:settingtraindarts}
\end{table}

\subsection{Optimal Values for the A$^2$M \texorpdfstring{$\rho_{\alpha}$}{} Hyperparameter} 
\label{appendix:rho}
\begin{figure}[H]
    \centering
    \includegraphics[width=0.51\columnwidth]{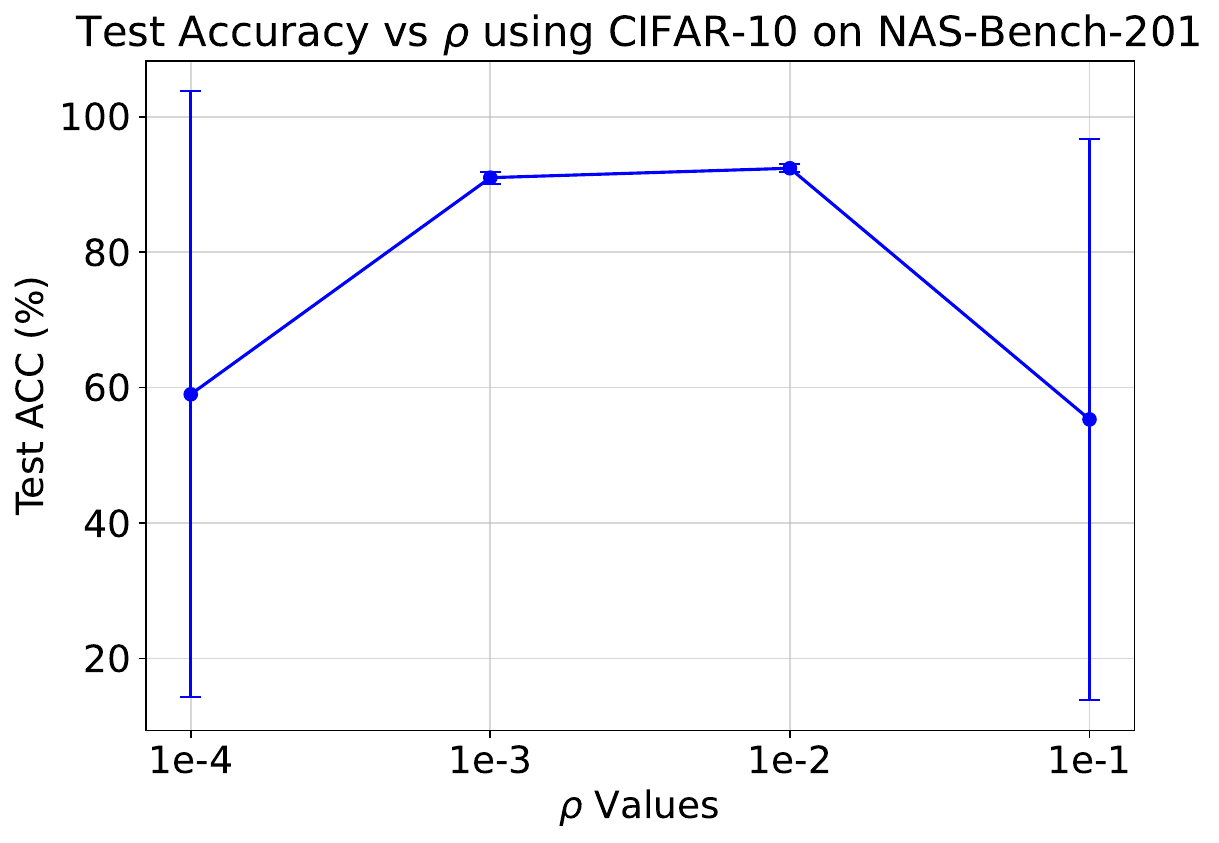} 
    \includegraphics[width=0.48\columnwidth]{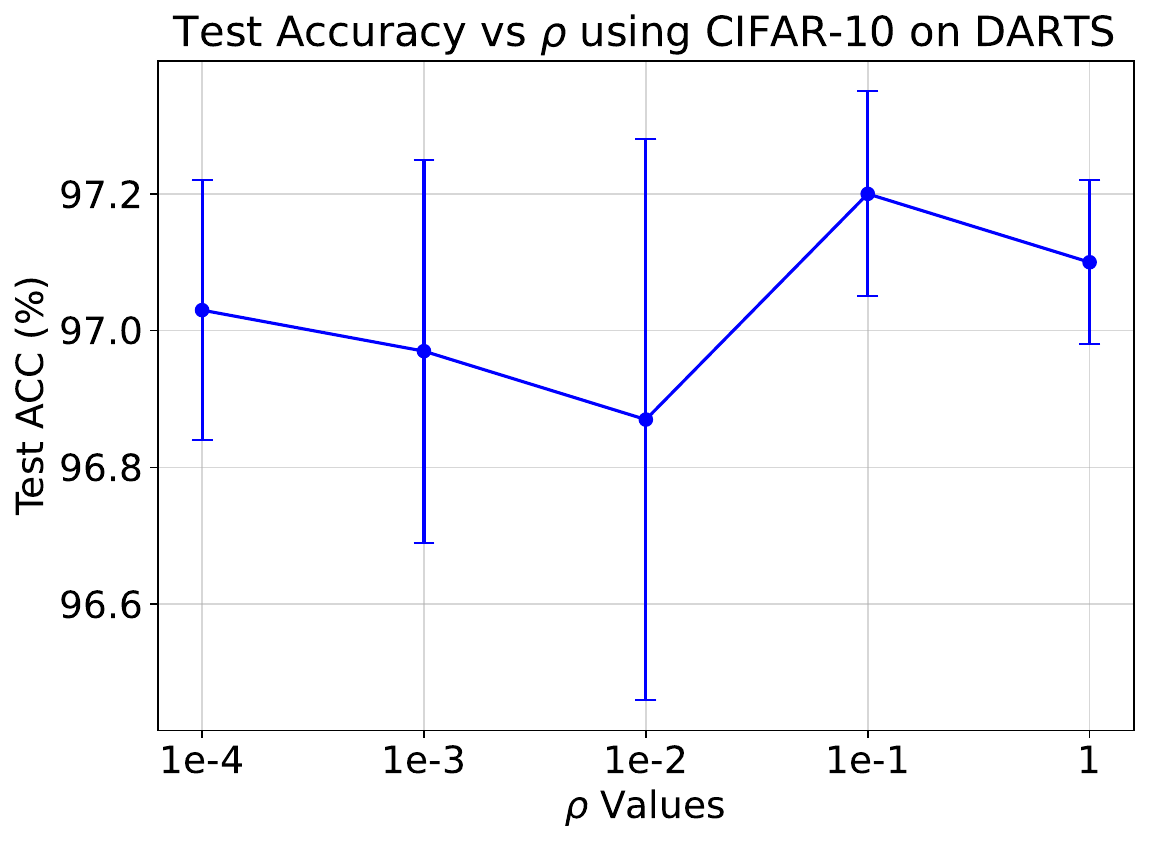} 
    \caption{Test accuracies of the final NN models found on CIFAR-10 on the NAS-Bench-201 (left) and DARTS (right) search spaces by A$^2$-DARTS in function of the hyperparameter~$\rho_{\alpha}$. \textcolor{black}{Error bars indicate the standard deviation over five runs with different random seeds.}
    }
    \label{fig:testrhonasbench}
    \label{fig:testrhodarts}
\end{figure}

This section outlines the optimal values of the $\rho_{\alpha}$ hyperparameter—introduced by our A$^2$M method—used in our experiments.
In Fig.~\ref{fig:testrhonasbench}, we report the mean and standard deviation of the final test accuracies on CIFAR-10, averaged over multiple runs, for different values of $\rho_{\alpha}$ within the A$^2$M framework. The best performance was observed for $\rho_{\alpha} = 10^{-2}$ on NAS-Bench-201 (92.39\% on average) and for $\rho_{\alpha} = 10^{-1}$ on DARTS (97.20\% on average).
We then use these values of $\rho_{\alpha}$ on all datasets.

\textcolor{black}{As shown in Fig.~\ref{fig:testrhodarts} (left), the performance of our method on NAS-Bench-201 is sensitive to the choice of $\rho_{\alpha}$. The error bars represent the standard deviation across five different runs, and they reveal a distinct pattern in search stability. We observe a high standard deviation for the extreme values of $\rho_{\alpha}$ ($10^{-4}$ and $10^{-1}$) but significantly lower variance for the intermediate values ($10^{-3}$ and $10^{-2}$). This indicates a strong sensitivity to the initial seed at the extremes, suggesting the method struggles to consistently converge to high-performing models at these values.
We can interpret this behavior as follows. A very low $\rho_{\alpha}$ ($10^{-4}$) makes A$^2$M behave similarly to standard first-order DARTS, which is known to exhibit high variance and instability on NAS-Bench-201. Conversely, a very high $\rho_{\alpha}$ ($10^{-1}$) appears to introduce noise by giving excessive weight to poorly-performing models from distant regions of the search space, disrupting convergence. The lower variance for $\rho_{\alpha}$ values of $10^{-3}$ and $10^{-2}$ suggests these values strike an effective balance, enabling stable convergence. These findings emphasize the importance of selecting an appropriate $\rho_{\alpha}$ value to ensure robust performance (confirming anyway that it is stable over two order of magnitude).}

\subsection{Data Augmentation on the Validation Set of NAS-Bench-201}
\label{appendix:dataaug}
DARTS-based methods typically include data augmentation applied to the validation set. It is important to notice that architecture parameters in DARTS-based methods are trained using the validation set, meaning that augmentation can significantly influence search dynamics.
Interestingly, as can be seen from Tables~\ref{tab:test}~and~\ref{tab:nasbench_valid}, we found that without data augmentation on NAS-Bench-201, standard DARTS does not suffer at every run of convergence to the degenerate architecture composed entirely of skip connections, although the issue remains present. 
This suggests a correlation between data augmentation and the DARTS instability towards degenerate solutions.
In contrast, other DARTS-based methods generally benefit from data augmentation, likely because their regularization mechanisms are themselves able to mitigate the skip connection issue.

\subsection{Implementing the $\beta$-DARTS regularization on the DARTS search space}
\label{betaweights}

The $\beta$-DARTS update rule has to be redefined to be compliant with the DARTS search space since this space contains two types of cells, i.e. normal and reduction ones~\cite{lambdadarts}. We could not find, as of the beginning of March 2025, in the official $\beta$-DARTS GitHub repository~\url{https://github.com/Sunshine-Ye/Beta-DARTS} scripts and code for directly running experiments with $\beta$-DARTS on the DARTS search space - in particular we could not find it in the \texttt{\_backward\_step} at line 80 of \url{https://github.com/Sunshine-Ye/Beta-DARTS/blob/master/optimizers/darts/architect.py}.

We therefore proceed to define the $\beta$-DARTS loss for the DARTS search space as:
\begin{equation}
\mathcal{L}_{\text{val}}\left(w, \alpha\right) + \lambda \left( w_{\text{nor}} \mathcal{L}_{\text{Beta}}\left(\alpha_{\text{nor}}\right)
 + w_{\text{red}} \mathcal{L}_{\text{Beta}}\left(\alpha_{\text{red}}\right) \right)
 \label{eq:betaweights}
\end{equation}
with $\lambda = \text{epoch}/2$ as done in the original $\beta$-DARTS implementation.
In our experiments we use the $\mathcal{L}_{\text{Beta}}$ regularization on both the architecture parameters of the normal and reduction cells, and we add the weights $w_{\text{nor}}$ and $w_{\text{red}}$ in the loss Eq.~\eqref{eq:betaweights} to perform an additional hyperparameter tuning which is not reported in the original $\beta$-DARTS papers~\cite{betadarts, betadarts++}. To this aim we investigate how the hyperparameters $w_{\text{nor}}$ and $w_{\text{red}}$ impact the performance by applying more or less regularization to the normal or reduction cells.
We report in Table~\ref{tab:test} the best results on test accuracy we obtain for $\beta$-DARTS on each dataset by tuning these hyperparameters (which corresponds to $w_{\text{nor}}=1.0$ and $w_{\text{red}}=1.0$).

In Table~\ref{tab:betadarts} we show the test accuracy results obtained by $\beta$-DARTS on the DARTS search space and on CIFAR-10, CIFAR-100 and ImageNet-16-120, with different combinations of the $w_{\text{nor}}$ and $w_{\text{red}}$ weights, corresponding to: $\beta$-DARTS with equal importance on normal and reduction cells ($w_{\text{nor}}=1.0$, $w_{\text{red}}=1.0$), $\beta$-NOR with more importance on the normal cell ($w_{\text{nor}}=1.6$, $w_{\text{red}}=0.4$), $\beta$-RED with more importance on the reduction cell ($w_{\text{nor}}=0.4$, $w_{\text{red}}=1.6$). 

\begin{table}[h!]
\caption{Final test accuracy of the models obtained by $\beta$-DARTS on the DARTS search space with different weights $w_{\text{nor}}$ and $w_{\text{red}}$ on the regularization of the normal and reduction cells. From top to bottom: (a) equal importance on normal and reduction cells ($w_{\text{nor}}=1.0$, $w_{\text{red}}=1.0$), (b) $\beta$-NOR with more importance on the normal cell ($w_{\text{nor}}=1.6$, $w_{\text{red}}=0.4$), (c) $\beta$-RED with more importance on the reduction cell ($w_{\text{nor}}=0.4$, $w_{\text{red}}=1.6$).}
\label{tab:betadarts}
\vskip 0.15in
\begin{center}
\begin{small}
\begin{sc}
\scalebox{1.0}{
\begin{tabular}{cccc}
\toprule
\textbf{Model} & \textbf{CIFAR-10} & \textbf{CIFAR-100} &
\textbf{ImageNet-16-120}
\\
\midrule
$\beta$-DARTS & $\mathbf{96.83\pm0.15}$ & $\mathbf{81.85\pm0.68}$ & $\mathbf{53.92\pm0.47}$ \\
$\beta$-NOR & $96.82\pm0.30$ & $81.49\pm1.00$ & $48.47\pm0.53$  \\
$\beta$-RED & $96.60\pm0.16$ & $81.47\pm0.78$ & $51.87\pm0.43$  \\
\bottomrule
\end{tabular}
}
\end{sc}
\end{small}
\end{center}
\vskip -0.1in
\end{table}

\section{Additional results}

In this section, we report additional results and ablation studies.

\subsection{Neighborhoods at the Extended Radius 2 and 3}
\label{appendix:radii}

\begin{figure}[h!]
    \centering
    \includegraphics[width=0.325\columnwidth]{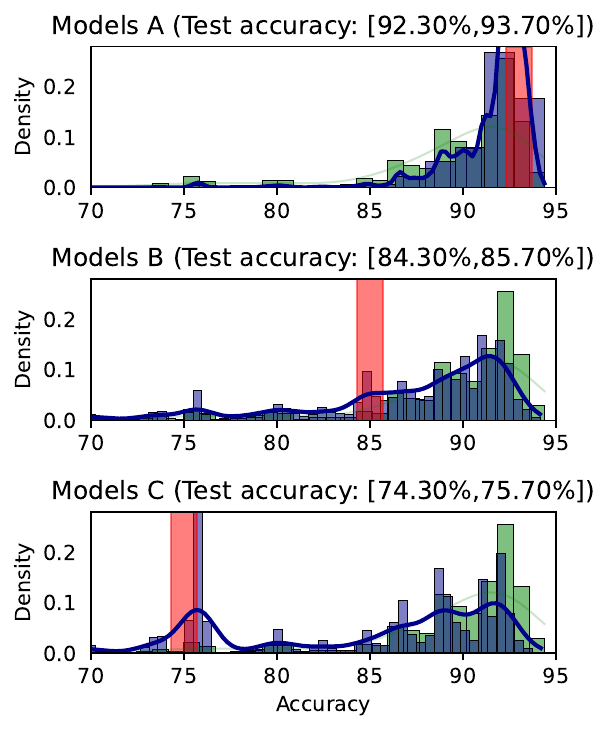} 
    \includegraphics[width=0.325\columnwidth]{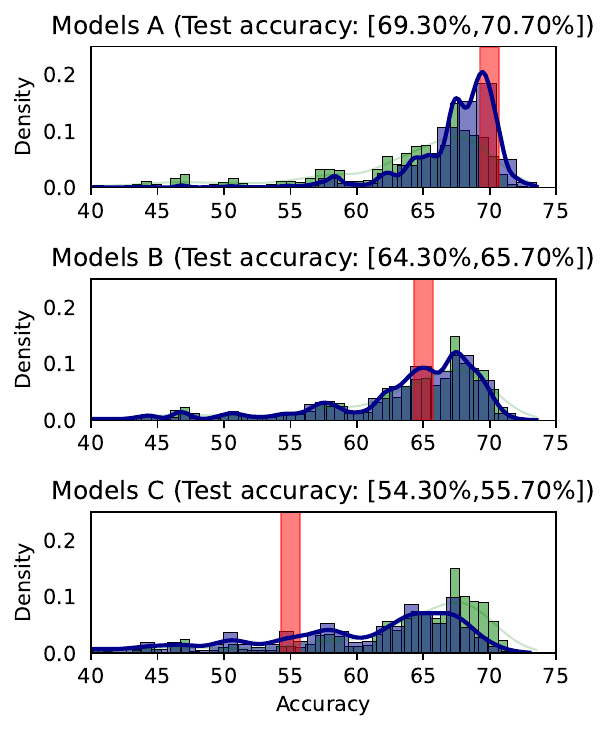} 
    \includegraphics[width=0.325\columnwidth]{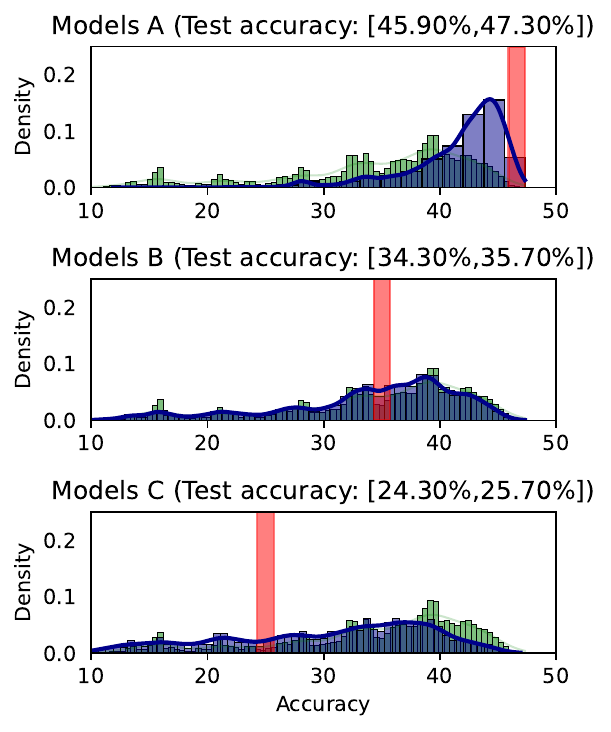} 
    \caption{
    Histogram of test accuracies on radius-2 neighborhoods on NAS-Bench-201 on CIFAR-10 (left column); CIFAR-100 (middle column); ImageNet-16-120 (right column), for different reference architectures in the search space. The shaded red area refers to the range of test accuracies of the reference architectures, that is also reported in each subplot title. For each dataset, three accuracy ranges (corresponding to high, medium, and low performance) have been identified according to the difficulty of the dataset. The blue distribution regards the neighboring architectures. The green distribution regards the whole search space.
    }
    \label{fig:nasbench-cifar10-radius2}
\end{figure}

\begin{figure}[h!]
    \centering
    \includegraphics[width=0.325\columnwidth]{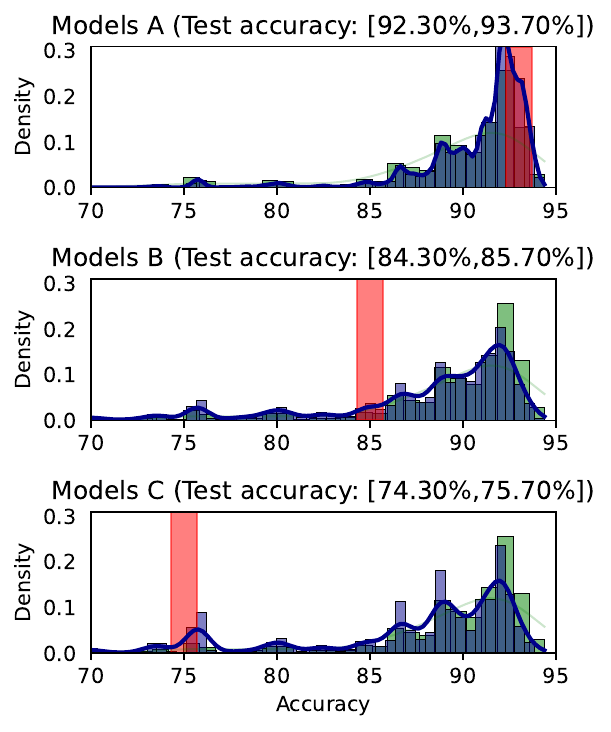} 
    \includegraphics[width=0.325\columnwidth]{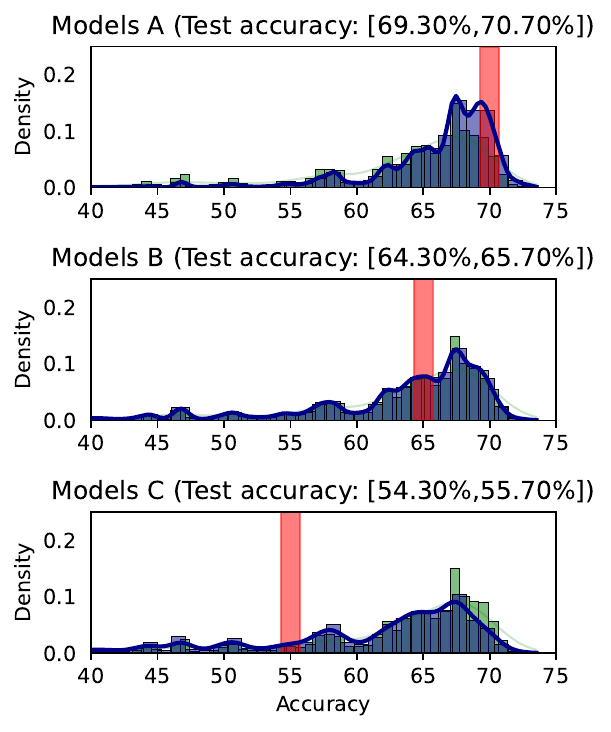} 
    \includegraphics[width=0.325\columnwidth]{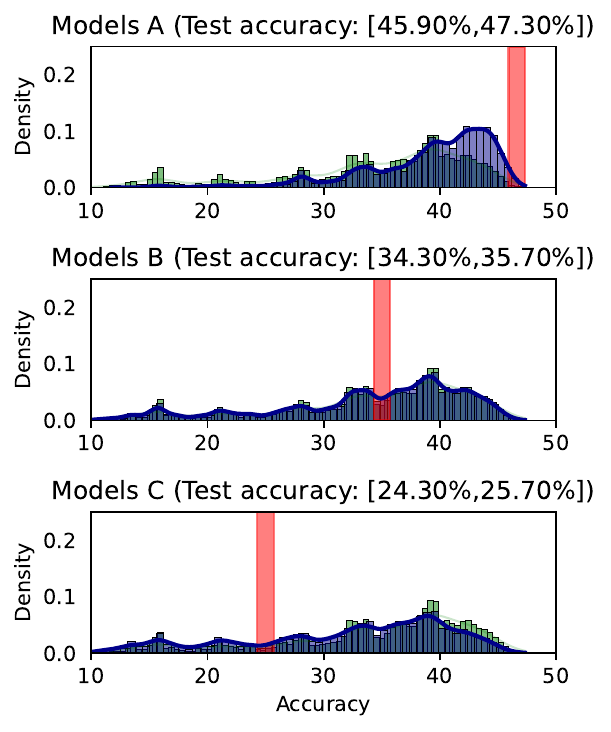} 
    \caption{Histogram of test accuracies on radius-3 neighborhoods on NAS-Bench-201 on (left column) CIFAR-10; (middle column) CIFAR-100; (right column) ImageNet-16-120, for different reference architectures in the search space. The red shaded area refers to the range of test accuracies of the reference architectures, that is also reported in each subplot title. For each dataset, three accuracy ranges (corresponding to high, medium, and low performance) have been identified according to the difficulty of the dataset.The blue distribution regards the neighboring architectures. The green distribution regards the whole search space. 
    }
    \label{fig:nasbench-cifar10-radius3}
\end{figure}

We report test accuracy histograms computed respectively over neighborhoods at radius $2$ and $3$ in Figs.~\ref{fig:nasbench-cifar10-radius2} and \ref{fig:nasbench-cifar10-radius3} for NAS-Bench-201.
Interestingly, we observe that in the Nas-Bench-201 case the accuracy distribution over the extended neighborhoods at radius~2 and~3 (blue) becomes similar to the distribution of the whole search space (green), due to the small size of the search space, and therefore it becomes impossible to observe architecture clustering. 

\begin{figure*}[t]
    \centering
    \includegraphics[width=0.60\columnwidth]{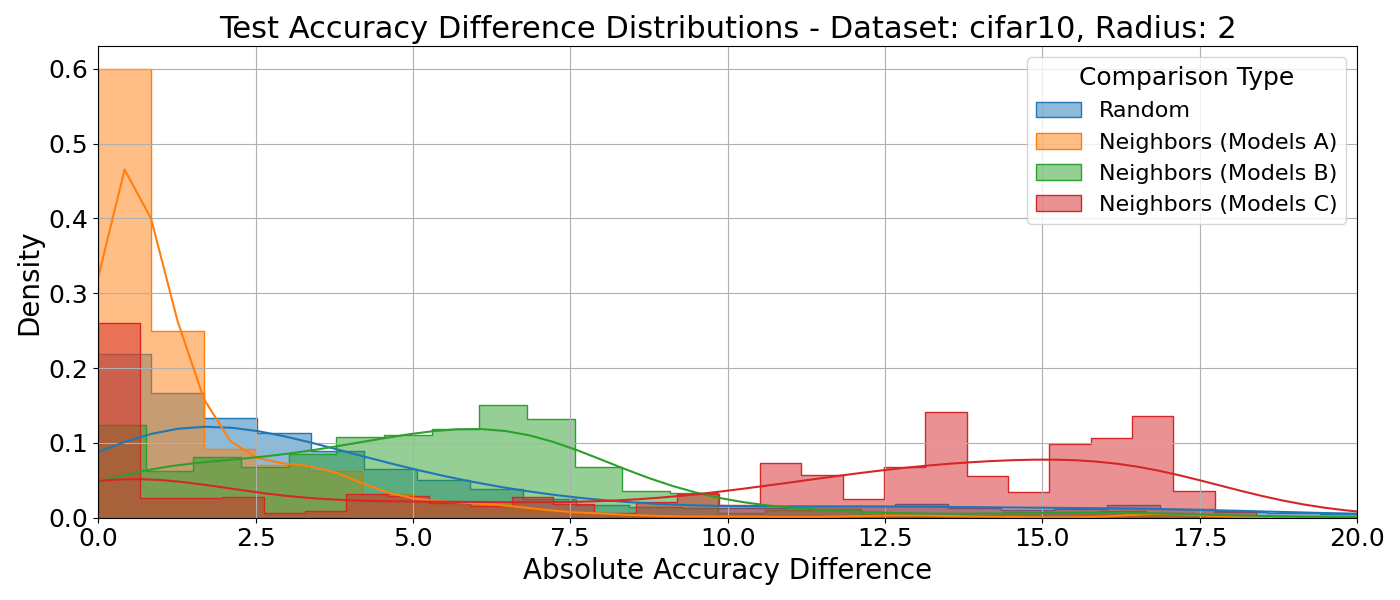} 
    \includegraphics[width=0.60\columnwidth]{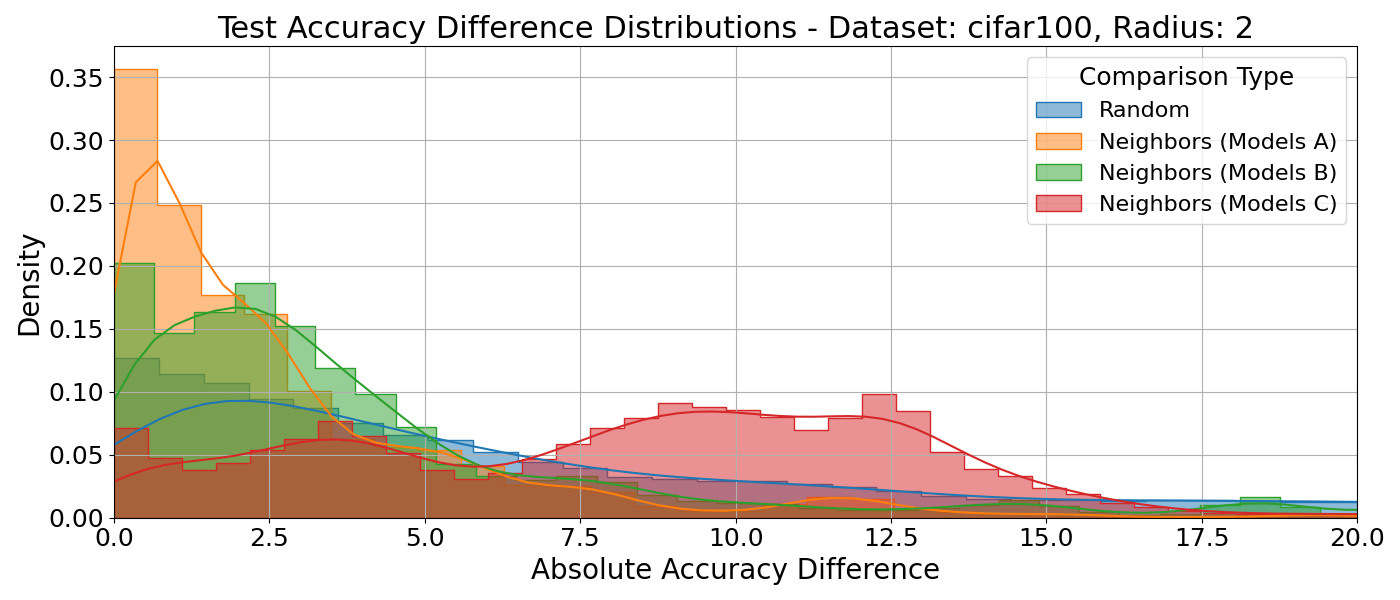} 
    \includegraphics[width=0.60\columnwidth]{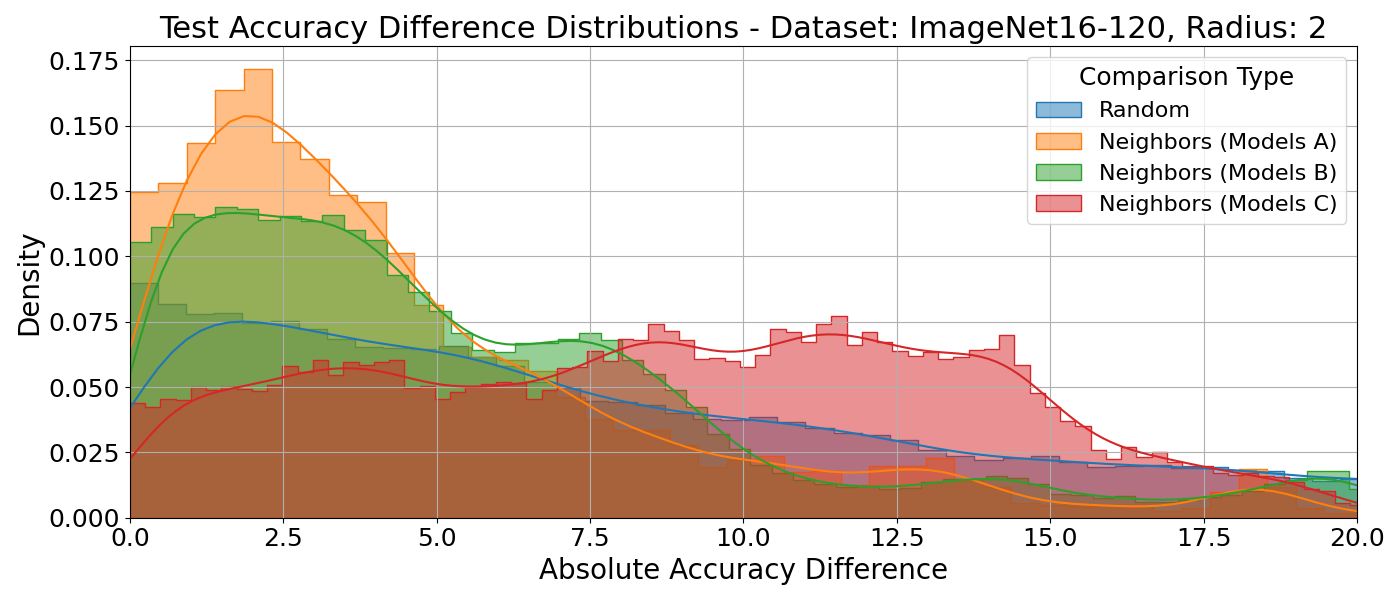} 
    \caption{
    Distributions of differences of test accuracies between random models in the search space and between reference architectures and their radius-2 neighborhoods on NAS-Bench-201 on CIFAR-10 (left), CIFAR-100 (middle), and ImageNet-16-120 (right), for different accuracy ranges of reference architectures in the search space. For each dataset, three accuracy ranges for the reference architecture were identified (corresponding to high, medium, and low performance), according to the difficulty of the dataset. The blue distribution refers to the random models, while the other ones refer to the reference architectures and their neighborhoods.
    The higher density of the orange distribution reveals the clustering property, being that models with similar accuracies tend to geometrically cluster together.
    }
    \label{fig:diff_cluster_radius2}
\end{figure*}

\begin{figure*}[t]
    \centering
    \includegraphics[width=0.60\columnwidth]{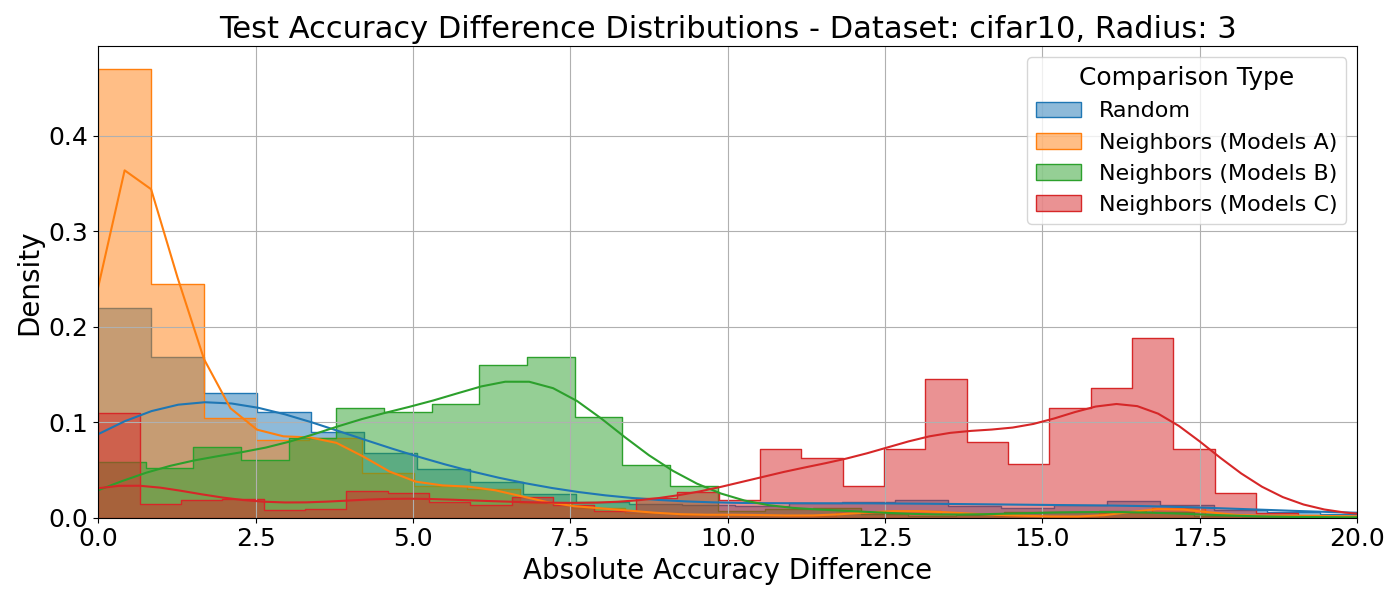} 
    \includegraphics[width=0.60\columnwidth]{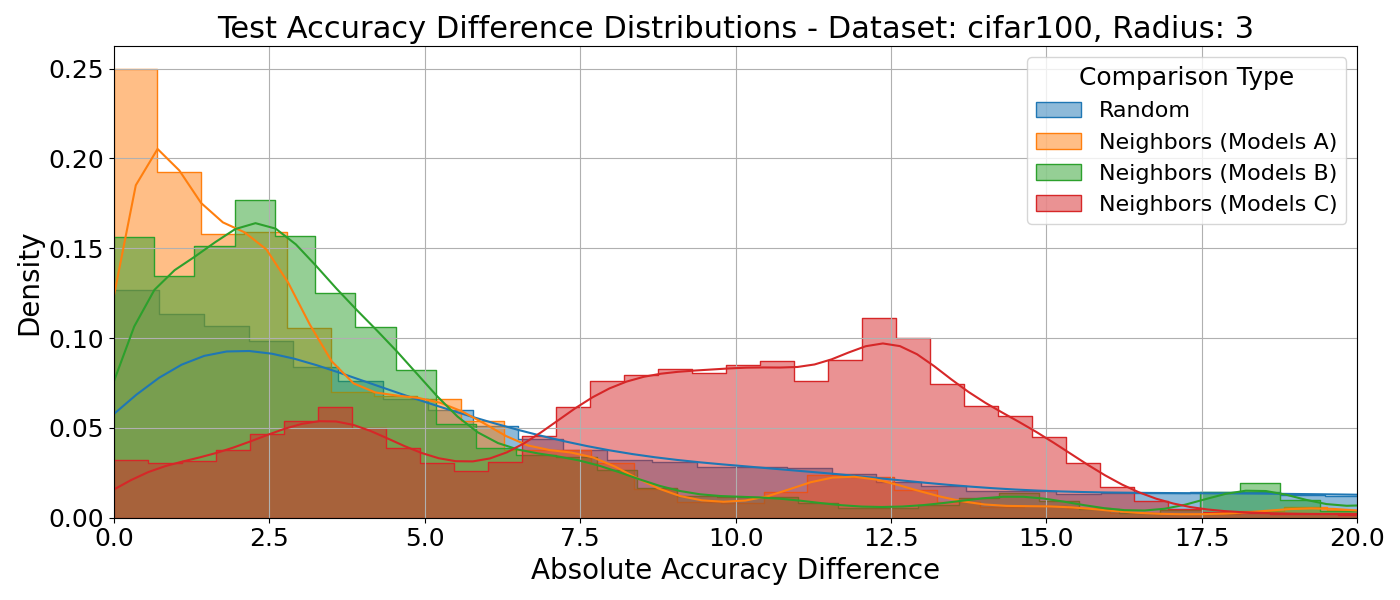} 
    \includegraphics[width=0.60\columnwidth]{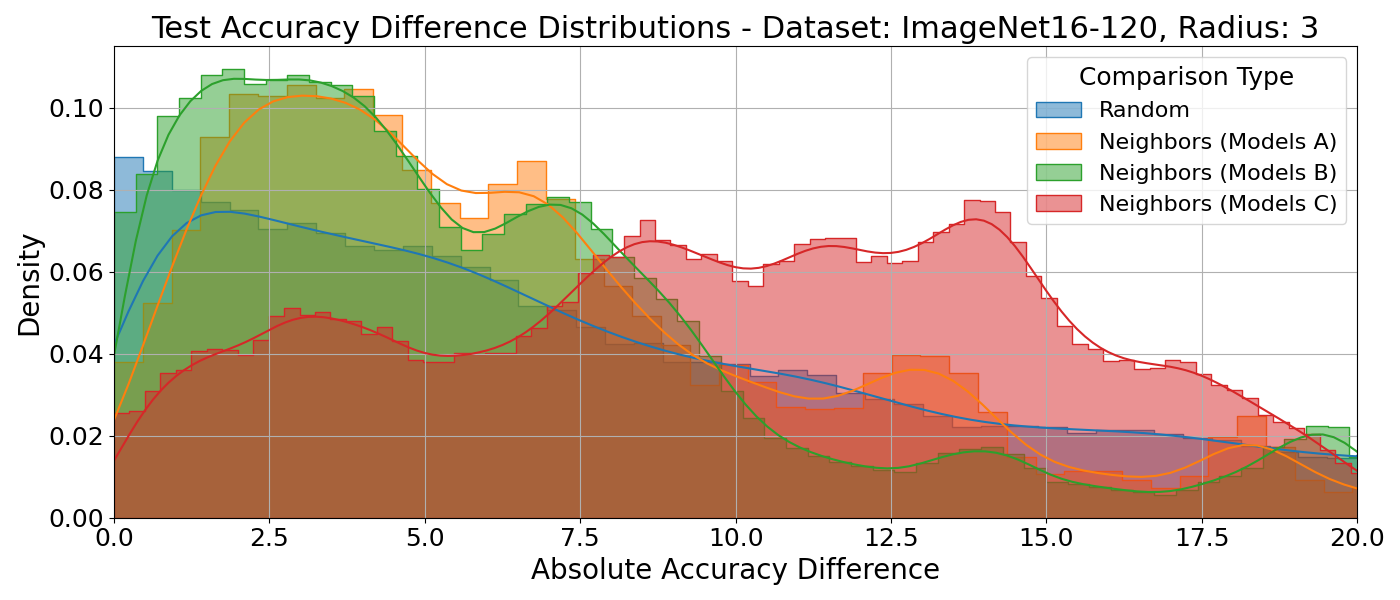} 
    \caption{
    Distributions of differences of test accuracies between random models in the search space and between reference architectures and their radius-3 neighborhoods on NAS-Bench-201 on CIFAR-10 (left), CIFAR-100 (middle), and ImageNet-16-120 (right), for different accuracy ranges of reference architectures in the search space. For each dataset, three accuracy ranges for the reference architecture were identified (corresponding to high, medium, and low performance), according to the difficulty of the dataset. The blue distribution refers to the random models, while the other ones refer to the reference architectures and their neighborhoods.
    The higher density of the orange distribution reveals the clustering property, being that models with similar accuracies tend to geometrically cluster together.
    }
    \label{fig:diff_cluster_radius3}
\end{figure*}

We can quantify the clustering property at the extended radius 2 and 3 by showing the distributions of differences of test accuracy between random models and models with high-, medium-, and low-accuracy and their neighborhoods at radius 2 and 3, reported in Figs.~\ref{fig:diff_cluster_radius2}~and~\ref{fig:diff_cluster_radius3}. Interestingly, medium-performing architectures seem to have slightly more similar radius-3 neighbors than high-performing architectures. This may suggest that the clustering property is limited up to a radius of 3 for ImageNet-16-120. It also means that even high-performing architectures suffer a drop in accuracy when perturbed by 3 or more layers on average.

\subsection{Validation accuracy on NAS-Bench-201}
\label{appendix:additional}

We report the accuracy of DARTS-based methods investigated in this paper on Nas-Bench-201 in Table~\ref{tab:nasbench_valid}.

\begin{table}[h!]
    \caption{Validation accuracy on NAS-Bench-201 search space across CIFAR-10, CIFAR-100, and ImageNet-16-120 datasets and several DARTS-based methods, showing standard and A$^2$-enhanced results. *DARTS and A$^2$-DARTS have been run without data augmentation on the valid set.}
    \label{tab:nasbench_valid}
    \vskip 0.15in
    \begin{center}
    \begin{small}
    \begin{sc}
    \scalebox{0.85}{
    \begin{tabular}{ccccccc}
    \toprule
    \multirow{2}{*}{\textbf{Model}} & \multicolumn{2}{c}{\textbf{CIFAR-10 Valid Acc (\%)}} & \multicolumn{2}{c}{\textbf{CIFAR-100 Valid Acc (\%)}} & \multicolumn{2}{c}{\textbf{ImageNet-16-120 Valid Acc (\%)}} \\
    \cline{2-7}
     & \textbf{Standard} & \textbf{A$^2$-} & \textbf{Standard} & \textbf{A$^2$-} & \textbf{Standard} & \textbf{A$^2$-} \\
    \midrule
    DARTS ($1^{st}$)*~\cite{liu_darts_2019} &  $70.86\pm34.09$ & \textbf{89.64$\pm$0.65} & $49.42\pm27.42$ & \textbf{68.60$\pm$1.69} & $27.12\pm 15.58$ & \textbf{41.96$\pm$1.86} \\
    $\beta$-DARTS~\cite{betadarts} & \textbf{91.40$\pm$0.18} & \textbf{91.40$\pm$0.18} & \textbf{72.70$\pm$1.0} & \textbf{72.70$\pm$1.0} & \textbf{45.97$\pm$0.37} & \textbf{45.97$\pm$0.37} \\
    $\Lambda$-DARTS~\cite{lambdadarts} & $89.75\pm2.10$ & \textbf{91.11$\pm$0.53} & $69.81\pm3.54$ & \textbf{72.00$\pm$1.72} & $43.18\pm2.95$ & \textbf{45.01$\pm$1.64} \\
    DARTS-PT\cite{wang2021rethinking} & $85.30\pm0.0$ & \textbf{85.40$\pm$0.0} & $61.05\pm0.0$  & \textbf{63.28$\pm$0.0} & \textbf{35.18$\pm$0.0} & $33.63\pm0.0$ \\
    DARTS-\cite{darts-} & \textbf{90.20$\pm$0.47} & \textbf{90.20$\pm$0.47} & \textbf{70.71$\pm$1.60} & \textbf{70.71$\pm$1.60} & \textbf{40.78$\pm$1.58} &  \textbf{40.78$\pm$1.58} \\
    SDARTS-RS\cite{sdarts-adv} & $75.21\pm0.25$ & \textbf{82.53$\pm$0.16} & $47.51\pm0.39$ & \textbf{54.53$\pm$0.31} & \textbf{27.79$\pm$0.68} & $27.45\pm0.22$ \\
    PC-DARTS\cite{pc-darts} & $68.28\pm0.31$ & \textbf{89.07$\pm$0.20} & $38.57\pm0.88$ & \textbf{66.43$\pm$0.46} & $18.87\pm0.65$ & \textbf{40.13$\pm$0.82} \\
    Optimal~\cite{nasbench201} & \multicolumn{2}{c}{\textbf{91.61}} & \multicolumn{2}{c}{\textbf{73.49}} & \multicolumn{2}{c}{\textbf{45.56}} \\
    \bottomrule
    \end{tabular}
    }
    \end{sc}
    \end{small}
    \end{center}
    \vskip -0.1in
\end{table}

\subsection{Statistical Analysis of Flatness}
\label{statanalysis}

We conducted a statistical analysis of the flatness by applying a Kolmogorov-Smirnov test on the distributions of neighbors at radius 1,2,3 of models A,B,C with respect to the total distribution of the NAS-Bench-201 search space, shown in Fig.~\ref{fig:nasbench-histograms-radius1}.
The results of this analysis is shown in Table~\ref{tab:ks_all_datasets}.
\begin{table*}[h!]
    \caption{Kolmogorov–Smirnov two-sample test results (D-statistic and sample size $n_2$) for NAS-Bench-201 across CIFAR-10, CIFAR-100, and ImageNet16-120. All tests used $n_1 = 15625$ (entire search space) and yielded p-values $\ll 0.001$ (two-sided). Models A, B, C refer to high, medium, low performance tiers.}
    \label{tab:ks_all_datasets}
\vskip 0.15in
\begin{center}
\begin{small}
\begin{sc}
\scalebox{0.85}{
\begin{tabular}{cc|cc|cc|cc}
\toprule
\makecell{\textbf{Models} \\ \textbf{Interval}} & \textbf{Radius} & 
\multicolumn{2}{c|}{\textbf{CIFAR-10}} & 
\multicolumn{2}{c|}{\textbf{CIFAR-100}} & 
\multicolumn{2}{c}{\textbf{ImageNet-16-120}} \\
\cmidrule{3-8}
 & & $D$ & $n_2$ & $D$ & $n_2$ & $D$ & $n_2$ \\
\midrule
A & 1 & 0.408 & 76056 & 0.472 & 35976 & 0.702 & 1464 \\
B & 1 & 0.370 & 8544  & 0.113 & 38976 & 0.146 & 15744 \\
C & 1 & 0.511 & 5496  & 0.366 & 6048  & 0.359 & 3672 \\
A & 2 & 0.267 & 760560 & 0.307 & 359760 & 0.484 & 14640 \\
B & 2 & 0.215 & 85440 & 0.040 & 389760 & 0.066 & 157440 \\
C & 2 & 0.318 & 54960 & 0.224 & 60480 & 0.214 & 36720 \\
A & 3 & 0.160 & 4056320 & 0.184 & 1918720 & 0.280 & 78080 \\
B & 3 & 0.127 & 455680 & 0.021 & 2078720 & 0.031 & 839680 \\
C & 3 & 0.183 & 293120 & 0.125 & 322560 & 0.119 & 195840 \\
\bottomrule
\end{tabular}
}
\end{sc}
\end{small}
\end{center}
\vskip -0.1in
\end{table*}

\end{document}